\documentclass[oneside,a4paper,11pt]{article}

\usepackage{natbib}
\usepackage{subcaption}
\usepackage{amsthm}
\usepackage{mathtools}
\usepackage[colorlinks]{hyperref}
\hypersetup{linkcolor=black}
\hypersetup{urlcolor=black}
\hypersetup{citecolor=black}
\usepackage{cleveref}
\usepackage{quiver}
\usepackage{tikz}
\usetikzlibrary{cd}
\usepackage{tabularx} 
\usepackage{booktabs}
\usepackage{multirow}
\usepackage{makecell}
\usepackage{arydshln}
\usepackage{subfiles}
\usepackage{acro}
\usepackage{babel}

\DeclareAcronym{ct}{
  short=CT,
  long=category theory,
}
\usepackage{rotating}

\usepackage{newpxtext,newpxmath}
\usepackage{geometry}

\begin{document}

\newtheorem{definition}{Definition}
\newtheorem{lemma}{Lemma}[section]

\title{Types of Relations: Defining Analogies with Category Theory}

\author{
  Claire Ott \hspace{8em} Frank Jäkel\\
\small{claire\_tabea.ott@tu-darmstadt.de\hspace{1em} jaekel@psychologie.tu-darmstadt.de}\\
\\
      Technische Universität Darmstadt}
\date{}
\providecommand{\keywords}[1]
{	
  \textbf{\textit{Keywords---}} #1
}

\newcommand{\C}{\mathcal{C}}
\newcommand{\D}{\mathcal{D}}
\newcommand{\M}{\mathcal{M}}
\newcommand{\sun}{\mathcal{S}}
\newcommand{\atom}{\mathcal{A}}
\newcommand{\water}{\mathcal{W}}
\newcommand{\heat}{\mathcal{H}}
\newcommand{\set}{\mathbf{Set}}
\newcommand{\cat}{\mathbf{Cat}}
\newcommand{\Hom}{\mathrm{Mor}}
\newcommand{\Ob}{\mathrm{Ob}}
\newcommand{\BE}{\dot{B}}
\newcommand{\BEE}{\ddot{B}}
\newcommand{\2}{\cdot,\cdot}
\newcommand{\1}{\cdot}
\def\te#1{\text{\emph{#1}}}

\definecolor{mygray}{gray}{0.9}
\maketitle

\begin{abstract}
In order to behave intelligently both humans and machines have to represent their knowledge adequately for how it is used.
Humans often use analogies to transfer their knowledge to new domains, or help others with this transfer via explanations. Hence, an important question is: What representation can be used to construct, find, and evaluate analogies? 
In this paper, we study features of a domain that are important for constructing analogies. We do so by formalizing knowledge domains as categories. We use the well-known example of the analogy between the solar system and the hydrogen atom to demonstrate how to construct domain categories. We also show how functors, pullbacks, and pushouts can be used to define an analogy, describe its core and a corresponding blend of the underlying domains. 
\end{abstract}


\section{Introduction}

Our ability to form analogies is central for cognition and communication \citep{Winston_1980}. We use analogies to transfer knowledge from one domain of knowledge to another, and in doing so we gain new insights and ideas. We also use analogies to solve novel problems. We routinely use analogies in teaching and learning, where we describe a novel domain as similar in certain aspects to another, familiar domain. Learning by analogy is widely believed to be one of the magic ingredients of human intelligence \citep{Gentner_2003}. Hence, if we want to build machines that learn and think like people \citep{Lake_etal_2017}, formal theories of analogy-making will likely play a major role in this enterprise \citep{Mitchell_2021}. Here, we present a formal definition of analogies as functors in category theory.

Cognitive scientists and researchers in AI have studied analogies for a long time and there are several approaches to formalize analogies. 
They share that an analogy is always formed between two domains -- a base domain and a target domain -- and consists of a mapping between the two that describes their (structural) similarities. The mapping makes it possible to transfer knowledge from the base domain to the target domain.
Most approaches agree that analogies involve structure alignment \citep{Gentner_Hoyos_2017}. Among those, Structure Mapping Theory \citep{Gentner_1983} has been the most influential and has been implemented in the Structure Mapping Engine \citep{Falkenhainer_etal_1989,Yan_Forbus_Gentner_2003,Forbus_etal_2017b}. Approaches differ, however, in how they represent the knowledge in the domains that underlie an analogy. Prominent mathematical formalisms are graphs \citep{Winston_1980}, algebraic models \citep{Dastani_Indurkhya_Scha_2000}, and first-order or second-order logic. Among the approaches that use logic, Heuristic-Driven Theory Projection is particularly well developed \citep{Schwering_Krumnack_Kühnberger_Gust_2009,Besold2013AnalogyEI}. Interestingly, the same formalization can be used for analogy-making and conceptual blending of two domains \citep{Eppe_Maclean_Confalonieri_Kutz_Schorlemmer_Plaza_Kühnberger_2018,Guhe_Pease_Smaill_Martinez_Schmidt_Gust_Kühnberger_Krumnack_2011}.  

Making an analogy is tantamount to discovering that two domains have the same underlying structure despite looking different superficially. As category theory is the branch of mathematics that studies abstract structures, several researchers have used category theory in their analyses of analogies. For example, \cite{Arzi-Gonczarowski_1999} model components of a camera and the human eye with the category of so-called \emph{perceptions} and a pullback of these two objects is used to find a new abstract object with their commonalities. 
\cite{Abdel-Fattah_Kuehnberger} model the syntax and semantics of analogies and use the category of models and structure preserving relations, while \cite{Navarrete_Dartnell_2017} use an algebraic model for analogy together with commutative diagrams and coequalizers to model analogies.

In this paper, we follow the classical structure mapping approach \citep{Gentner_1983} which emphasizes the relations between entities in a domain. For simple analogies, knowledge domains can be represented as graphs with entities as nodes and binary relations as edges in the graph. The category of graphs can then be used to analyze possible relations and blends between such graphs \citep{ott_2025}. However, these graphs cannot easily represent n-ary and higher-order relations, which are known to be impor\-tant for analogy-making \citep{Gentner_1983}. Therefore, here, we will represent each domain as its own category. In this way, we are able to have relations between more than two entities and we can build higher-order relations using objectification. As in other approaches, a domain category consists of entities and predicates that encode the relevant knowledge about these entities.
Importantly, objectification of relations makes it straightforward to use types to identify the structural role each entity and predicate plays within a domain.

As a running example we illustrate our approach with the analogy between the solar system and the Bohr model of a hydrogen atom. Entities in the solar system domain are the \emph{sun} and planets \emph{venus} and \emph{mars}. The atom domain has a \emph{nucleus} and an \emph{electron} as entities. Relations are, for example, \emph{attracts(sun, mars)} or \emph{revolves around(electron, nucleus)}. The transfer in this classic example \citep{Gentner_1983} is that the electron in a hydrogen atom revolves around the nucleus like planets revolve around the sun.

In the following section, we will first describe how to build a domain category using the example of the solar system. We will also introduce all necessary concepts from \ac{ct}. We then define analogies as functors in \autoref{sec:analogy} and look at how pullbacks can be used to form a new category that captures the core of an analogy and how pushouts can be used to form a blend of two domains.

\section{Formalizing Domains as Categories}\label{sec:basic_cat}

\ac{ct} is the study of mathematical structures. A category consists of a class of objects and morphisms between these objects. These morphisms can be many different kinds of maps, and the rules of \ac{ct} only define how they act together, not their specific properties. The concept of a category is therefore very abstract, and thus applicable in many different areas of mathematics. Here, we use categories to represent the knowledge in the domains that underlie an analogy. The representations that we will construct as domain categories share many similarities with more common formalisms, like first- and higher-order logics, but focus more strongly on relations between entities.

\subsection{Introduction to Categories}
Before we can define domain categories and give a concrete example with the domain of the solar system, we will briefly introduce all necessary concepts from \ac{ct}. For a more in-depth introduction to \ac{ct} readers should consult a textbook on the topic \citep[e.g.,][]{David_Spivak_2014,Awodey_2010}.

\begin{definition}[Category]
    A category $\C$ consists of a class of objects $\Ob(\C)$ and a set $\Hom_\C(X,Y)$ of morphisms from $X$ to $Y$ for each pair of objects $X,Y\in \Ob(\C)$.
    For every object $X\in \Ob(\C)$ there exists an identity morphism $id_X\in \Hom_\C(X,X)$. There is also a composition function $\circ\colon\Hom_\C(Y,Z)\times \Hom_\C(X,Y)\rightarrow \Hom_\C(X,Z) $ for every three objects $X,Y,Z\in \Ob(\C)$.

    The following rules must hold in every category:
    \begin{enumerate}
        \item $f\circ id_X = f$ and $id_Y\circ f = f$ for every $X,Y\in \Ob(\C)$ and $f\in \Hom_\C(X,Y)$.
        \item $(h\circ g)\circ f = h\circ(g\circ f)$ for all $f\colon W\rightarrow X$, $g\colon X\rightarrow Y$ and $h\colon Y\rightarrow Z$ with $W,X,Y,Z\in \Ob(\C)$.
    \end{enumerate}
    
\end{definition}
We omit indices if they are clear from the context. We write $f\colon X\rightarrow Y$ for a morphism from $X$ to $Y$, where $X$ is the domain of $f$ and $Y$ the codomain.
A class can be a set or collection of sets, however in the remainder of this paper we will only consider categories with proper sets of objects. Such categories are called small categories.

Sometimes only part of a category is relevant for a certain analysis or it is interesting to consider a category as part of a bigger one. In these situations we use subcategories. 
\begin{definition}[Subcategory]\label{def:subcategory}
    A subcategory $\C'$ of a category $\C$ has objects $\Ob(\C')\subseteq \Ob(\C)$ and morphisms $\Hom_{\C'}(X,Y) \subseteq \Hom_\C(X,Y)$ for all $X,Y\in \Ob(\C')$.
    A subcategory has to contain the morphisms $id_X$ for all $X\in \Ob(\C')$ and for all $f\colon X\rightarrow Y$ and $g\colon Y\rightarrow Z$ in $\Hom_{\C'}$, the composition $g\circ f$ has to be in $\Hom_{\C'}$ so it is itself a category.
\end{definition}

We formalize every domain as its own category and will now describe how such a category is constructed. A domain category has base-types as objects which are different sets. \cite{Gentner_1983} describes domains as consisting of objects, attributes of objects and relations between objects. To avoid confusion with the term \textit{object} from \ac{ct}, we will refer to the basic objects in a domain as \emph{entities}. Every domain has a set of entities $E$. This set is one object in the domain category. For our example, we build the base-domain $\sun$ of the solar system with the object $E = \{\te{sun}, \te{mars}, \te{venus}\}$. Similarly, the target-domain $\atom$ of the hydrogen atom has an object $E = \{\te{nucleus}, \te{electron}\}$. We will now show step-by-step how to construct a domain category by adding more objects and morphisms that represent various components and predicates of the domain. The goal is to encode all relevant knowledge in a structured way that helps to find possible analogies, while keeping the categories small to reduce the amount of false correspondences.

Attributes of entities are unary relations and can be represented in a domain category as morphisms that map every entity to a boolean value. Hence, we require the category of a domain to also contain a set $B$ of boolean values and a morphism from $E$ to $B$ for each attribute. For example, in $\sun$ there is an object $B = \{\te{true}, \te{false}\}$ and the set of morphisms $\Hom(E,B)$ from $E$ to $B$ includes $\te{hot}(\1)$ which maps $\te{sun}$ to $\te{true}$ and the planets to $\te{false}$.

We must note here that there can be several attributes that behave identically, that is they map entities to the same boolean values. For example, there might be an attribute $\te{yellow}(\1)$ that also maps the $\te{sun}$ to $\te{true}$ and the planets to $\te{false}$. We want to discern between such morphisms. Hence, these morphisms are not simply functions. We think of them as named functions where $f(x) = g(x)$ for all $x$ does not imply $f = g$.

Another object of all domain categories, as we define them, is a unit object that has a similar role as $B$. It makes it possible to represent the single entities, which are elements of objects, as morphisms.

\begin{definition}[Unit object]
   We call an object $1_\C$ in a category $\C$, where the objects can be identified with sets, unit object if for every object $X\in \Ob(\C)$ and every element $x\in X$ there exists exactly one morphism $x\colon 1_\C\rightarrow X$.  
\end{definition}

We require that every domain category $\C$ contains the set $1_\C = \{*\}$, which is a set containing exactly one element (the $*$) as a unit object. For every object $X\in \Ob(\C)$ and for all elements $x\in X$ there is one unique morphism in $\Hom_{\C}(1_\C,X)$, namely, $x\colon 1_\C\rightarrow X$, that maps $*$ to the specific element $x$. 

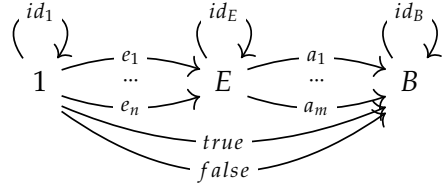
\begin{figure}
    \centering
\[\begin{tikzcd}[ampersand replacement=\&]
	1 \&\& E \&\& B
	\arrow["{id_1}"{description}, from=1-1, to=1-1, loop, in=55, out=125, distance=10mm]
	\arrow[""{name=0, anchor=center, inner sep=0}, "{e_1}"{description}, shift left, curve={height=-6pt}, from=1-1, to=1-3]
	\arrow[""{name=1, anchor=center, inner sep=0}, "{e_n}"{description}, shift right, curve={height=6pt}, from=1-1, to=1-3]
	\arrow["false"{description}, shift right=2, curve={height=30pt}, from=1-1, to=1-5]
	\arrow["true"{description}, shift right=2, curve={height=18pt}, from=1-1, to=1-5]
	\arrow["{id_E}"{description}, from=1-3, to=1-3, loop, in=55, out=125, distance=10mm]
	\arrow[""{name=2, anchor=center, inner sep=0}, "{a_m}"{description}, shift right, curve={height=6pt}, from=1-3, to=1-5]
	\arrow[""{name=3, anchor=center, inner sep=0}, "{a_1}"{description}, shift left, curve={height=-6pt}, from=1-3, to=1-5]
	\arrow["{id_B}"{description}, from=1-5, to=1-5, loop, in=55, out=125, distance=10mm]
	\arrow["\cdots"{description}, draw=none, from=0, to=1]
	\arrow["\cdots"{description}, draw=none, from=3, to=2]
\end{tikzcd}\]
    \caption{The basic building blocks of a domain category.}
    \label{fig:small-ex}
\end{figure}

So far, a domain category has the objects $E$, $B$ and $1$ and morphisms from $E$ to $B$ and from $1$ to $E$ and $B$. The composition of a morphism $e\colon 1\rightarrow E$ describing an entity and one from $E$ to $B$ describing an attribute is either $\te{true}\colon  1\rightarrow B$ or $\te{false}\colon 1\rightarrow B$ depending on whether that entity has the attribute or not. We equip all objects with an identity morphism $id_E$, $id_B$ and $id_1$ to complete the category as shown in \autoref{fig:small-ex}. From now on, we will usually omit identity morphisms, because they clutter tables and figures, but they are always part of a category. 

\subsection{Products: Binary and n-ary Relations}

So far, a domain category only contains unary relations. We will now look at products and how they can be used to include binary and n-ary relations.

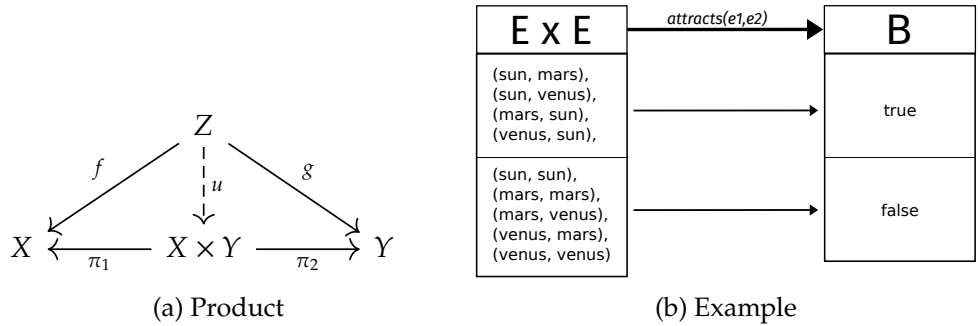
\begin{figure}[h]
    \centering
    \begin{subfigure}{0.4\textwidth}
\begin{tikzcd}[ampersand replacement=\&, sep=scriptsize]
	\&\& Z \\
	\\
	X \&\& {X\times Y} \&\& Y
	\arrow["{\pi_1}", from=3-3, to=3-1]
	\arrow["{\pi_2}"', from=3-3, to=3-5]
	\arrow["{f}"', from=1-3, to=3-1]
	\arrow["{g}", from=1-3, to=3-5]
	\arrow["u", dashed, from=1-3, to=3-3]
\end{tikzcd}
\caption{Product}
\label{fig:product}
\end{subfigure}
\begin{subfigure}{0.5\textwidth}
\centering
   \includegraphics[width = 0.9\textwidth]{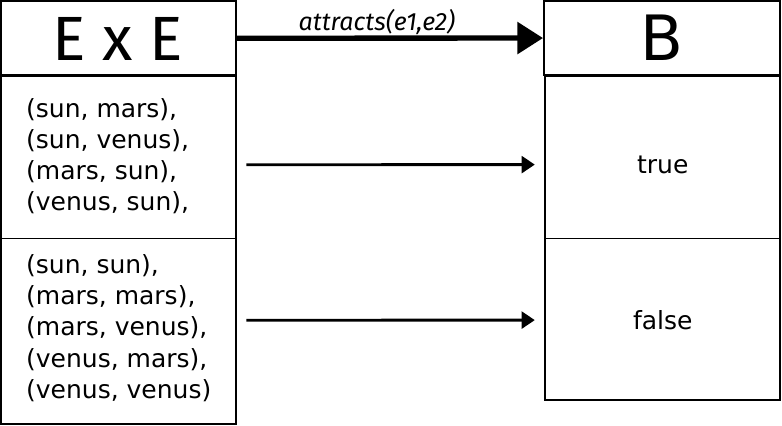}
   \caption{Example}
   \label{fig:ex-attracts}
\end{subfigure}
\caption{(a) Product of two objects $X$ and $Y$: For any object $Z$ and morphisms $f$ and $g$ there is a unique morphism $u$ such that the diagram commutes. Dashed arrows indicate unique morphisms. (b) Example of a morphism describing a binary relation: $\te{attracts}$ is a morphism that maps pairs of entities to \emph{true} or \emph{false}. }
\end{figure}

A binary relation maps a pair of entities to \emph{true} if they are in this relation and \emph{false} if they are not. So we need an object in the category that contains these pairs. This object is the Cartesian product of the set of entities with itself. It is also a product in the more general sense of \ac{ct} as shown in \autoref{fig:product}.

\begin{definition}[Product]\label{def:product}
    Let $\C$ be a category. A product of two objects $X,Y\in \Ob(\C)$ is an object $X\times Y\in \Ob(\C)$ with two projection morphisms $\pi_1\colon X\times Y \rightarrow X$ and $\pi_2\colon X\times Y \rightarrow Y$ such that for all objects $Z\in \Ob(\C)$ and morphisms $f\colon Z\rightarrow X$ and $g\colon Z\rightarrow Y$ there exists a unique $u\colon Z\rightarrow X\times Y$ such that $\pi_1\circ u = f$ and $\pi_2\circ u = g$. 
\end{definition}

We have defined three basic objects, $1$, $E$ and $B$ for a domain category. All other objects are composed of these three. We can add the product of two objects $X, Y\in \Ob(\C)$ to a domain category $\C$ with the following construction. The product is defined as the set $X\times Y = \{(x,y)|x\in X, y\in Y\}$. A morphism $(x,y)\colon 1_\C \rightarrow X\times Y$ is added for each element in $X\times Y$. Additionally, the projections $\pi_1(x,y) = x$ and $\pi_2(x,y) = y$ are added. For any object $Z\in Ob(\C)$ with morphisms $f\colon Z\rightarrow X$ and $g\colon Z\rightarrow Y$ the morphism $\langle f, g\rangle\colon Z \rightarrow X\times Y$ which is defined as the map $\langle f, g\rangle(z) = (f(z), g(z))$ is added to fulfill the property of a product. Additional morphisms can be added, for example to describe binary relations $r\colon E\times E\rightarrow B$, as long as the product properties are still fulfilled. Concatenations of any new morphism with other morphisms must also be added to adhere to the properties of a category, usually as $f\circ g = f(g(\1))$. If the domain of $g$ is $1_\C$, the concatenation is the corresponding existing morphism pointing to a specific element in the codomain of $f$. For example, $\te{hot}(\1)\circ\te{sun}(\1)= \te{true}(\1)$. If there is already a morphism $m$ describing this concatenation we can just say $f\circ g = m$. For example, there is the morphism $\langle id_E, id_E\rangle\colon E\rightarrow E\times E$ that maps an entity to a pair of this entity. We use $\delta$ as a shorthand for this morphism. Concatenating this morphism with one of the two projections $\pi\colon E\times E\rightarrow E$ yields the same entity. So $\pi_1\circ \langle id_E,id_E\rangle = id_E$. This does not work the other way round, so we have to add a morphism $\langle id_E,id_E\rangle\circ \pi_1\colon  E\times E\rightarrow E\times E$ that maps a pair of entities $(e_1,e_2)$ to a pair $(e_1,e_1)$. 

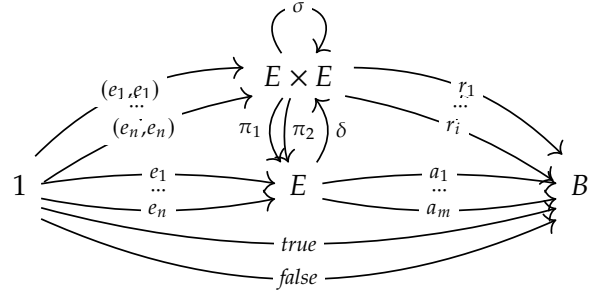
\begin{figure}
    \centering
\[\begin{tikzcd}[ampersand replacement=\&,row sep=2.25em]
	\&\&\& {E\times E} \\
	1 \&\&\& E \&\&\& B
	\arrow["\sigma"{description}, from=1-4, to=1-4, loop, in=55, out=125, distance=10mm]
	\arrow["{\pi_1}"', shift right, curve={height=6pt}, from=1-4, to=2-4]
	\arrow["{\pi_2}", curve={height=6pt}, from=1-4, to=2-4]
	\arrow[""{name=0, anchor=center, inner sep=0}, "{r_1}"{description}, shift left=2, curve={height=-12pt}, from=1-4, to=2-7]
	\arrow[""{name=1, anchor=center, inner sep=0}, "{r_i}"{description}, shift right, curve={height=-6pt}, from=1-4, to=2-7]
	\arrow[""{name=2, anchor=center, inner sep=0}, "{(e_1,e_1)}"{description}, shift left=2, curve={height=-12pt}, from=2-1, to=1-4]
	\arrow[""{name=3, anchor=center, inner sep=0}, "{(e_n,e_n)}"{description}, shift right, curve={height=-6pt}, from=2-1, to=1-4]
	\arrow[""{name=4, anchor=center, inner sep=0}, "{e_1}"{description}, curve={height=-6pt}, from=2-1, to=2-4]
	\arrow[""{name=5, anchor=center, inner sep=0}, "{e_n}"{description}, shift right, curve={height=6pt}, from=2-1, to=2-4]
	\arrow["{\te{false}}"{description}, shift right=3, curve={height=30pt}, from=2-1, to=2-7]
	\arrow["{\te{true}}"{description}, shift right=2, curve={height=18pt}, from=2-1, to=2-7]
	\arrow["\delta"', shift right, curve={height=6pt}, from=2-4, to=1-4]
	\arrow[""{name=6, anchor=center, inner sep=0}, "{a_m}"{description}, shift right, curve={height=6pt}, from=2-4, to=2-7]
	\arrow[""{name=7, anchor=center, inner sep=0}, "{a_1}"{description}, curve={height=-6pt}, from=2-4, to=2-7]
	\arrow["\cdots"{description}, shorten <=2pt, shorten >=2pt, Rightarrow, from=0, to=1]
	\arrow["\cdots"{description}, shorten <=2pt, shorten >=2pt, Rightarrow, from=2, to=3]
	\arrow["\cdots"{description}, draw=none, from=4, to=5]
	\arrow["\cdots"{description}, draw=none, from=7, to=6]
\end{tikzcd}\]
    \caption{A domain category with the object $E\times E$, equipped with morphisms describing attributes ($a$) and binary relations ($r$). }
    \label{fig:cat_EE}
\end{figure}

We can equip our domain category with some specific products. The object $E\times E$ in $\sun$ is the set of all pairs of entities and the projections $\pi_1$ and $\pi_2$ map each such pair to its first or second element. This means that binary relations can be expressed as morphisms $r\colon  E\times E \rightarrow B$. Using the above construction, it is also possible to add the product of $(E\times E) \times E$ for tertiary relations and iteratively more complex products with $E$ for n-ary relations. Every n-ary relation can be expressed as a morphism $r\colon  E\times \dots \times E \rightarrow B$. For easier readability, we will often write $X^n$ in instead of $X\times \dots \times X$. We will also use $\sigma\colon E\times E \rightarrow E\times E$ as shorthand for $\langle \pi_2,\pi_1\rangle$, which describes the permutation of the two entities in each pair in $E\times E$. One example binary relation in our category $\sun$ of the solar system is $\te{attracts}(\2)$ which maps every pair of the $\te{sun}$ and a planet to $\te{true}$ and all other pairs to $\te{false}$ as shown in \autoref{fig:ex-attracts}. 
\autoref{fig:cat_EE} shows the basic structure of a domain category with the product object $E\times E$. 

\subsection{Exponentials: Objectification of Relations}
\begin{figure}[h]
\centering
    \begin{subfigure}{0.45\textwidth}
\[\begin{tikzcd}[ampersand replacement=\&,column sep=scriptsize]
	{Y^X} \&\& {Y^X\times X} \&\& Y \\
	\\
	Z \&\& {Z\times X}
	\arrow["{\pi_1}"', color={rgb,255:red,153;green,153;blue,153}, from=1-3, to=1-1]
	\arrow["{\varepsilon_{X,Y}}", from=1-3, to=1-5]
	\arrow["{\lambda(f)}", dashed, from=3-1, to=1-1]
	\arrow["{\lambda(f)\times id_X}", from=3-3, to=1-3]
	\arrow["f"', from=3-3, to=1-5]
	\arrow["{\pi_1}", color={rgb,255:red,153;green,153;blue,153}, from=3-3, to=3-1]
\end{tikzcd}\]
    \caption{Exponential with a binary $f$}
    \label{fig:exponential}
\end{subfigure}
\begin{subfigure}{0.45\textwidth}
    \centering
\[\begin{tikzcd}[ampersand replacement=\&,column sep=scriptsize]
	{Y^X} \&\& {Y^X\times X} \&\& Y \\
	\\
	{Z} \&\& {Z\times X} \&\& \textcolor{rgb,255:red,153;green,153;blue,153}{X}
	\arrow["{\pi_1}"', color={rgb,255:red,153;green,153;blue,153}, from=1-3, to=1-1]
	\arrow["{\varepsilon_{X,Y}}", from=1-3, to=1-5]
	\arrow["{\lambda(g)}", dashed, from=3-1, to=1-1]
	\arrow["{\lambda(g)\times id_X}", from=3-3, to=1-3]
	\arrow["{g' = g\circ\pi_2}"{description}, from=3-3, to=1-5]
	\arrow["{\pi_1}", color={rgb,255:red,153;green,153;blue,153}, from=3-3, to=3-1]
	\arrow["{\pi_2}"', color={rgb,255:red,153;green,153;blue,153}, from=3-3, to=3-5]
	\arrow["g"', color={rgb,255:red,153;green,153;blue,153}, from=3-5, to=1-5]
\end{tikzcd}\]
    \caption{Exponential with a unary $g$}
    \label{fig:terminal-exponential}
\end{subfigure}
\caption{(a) An exponential consists of an object $Y^X$ and an evaluation morphism $\varepsilon_{X,Y}$ such that for all morphisms $f$ the above diagram commutes for a unique $\lambda(f)$. (b) The morphism $g\colon X \rightarrow Y$ must also be represented in $Y^X$ because it can be composed with a projection to a morphism form $Z\times X$ to $Y$ for some $Z\in \Ob(\C)$.}
\end{figure}
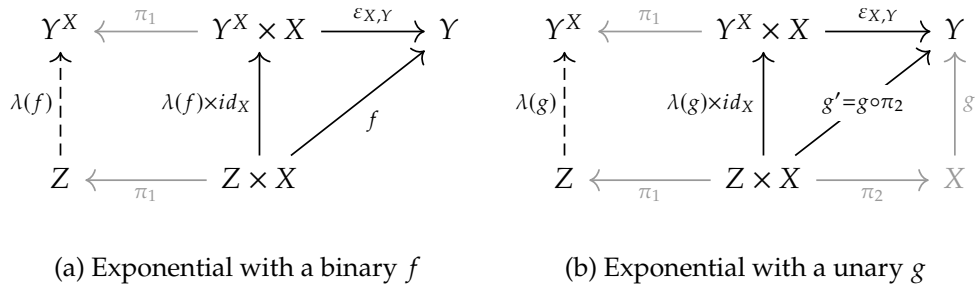

\cite{Gentner_1983} emphasizes the importance of relations on relations for building analo\-gies. Relations which are part of such higher-order predicates are more involved in the general structure of a domain and therefore more important for the transfer. To include higher-order predicates in a domain we need to objectify the existing relations. We use \emph{exponential objects} for this objectification. The definition is illustrated in \autoref{fig:exponential}.

\begin{definition}[Exponential]
The exponential of the objects $X,Y\in \Ob(\C)$ is an object $Y^X\in \Ob(\C)$ together with a morphism $\varepsilon_{X,Y}\colon  Y^X\times X \rightarrow Y$ such that for every morphism $f\colon Z\times X\rightarrow Y$ there is a unique morphism $\lambda(f)\colon Z \rightarrow Y^X$ such that $\varepsilon_{X,Y}\circ (\lambda(f)\times id_X) = f$.    
\end{definition}

In our domain categories, $Y^X$ will contain all the named functions from $X$ to $Y$. The product $Y^X\times X$ pairs all these functions with all elements of $X$. The morphism $\varepsilon_{X,Y}$ is called the evaluation morphism because it evaluates a function from $Y^X$ on an element from $X$. 
The definition of exponential object requires that every morphism $f\colon Z\times X\rightarrow Y$ which takes an element $z\in Z$ and $x\in X$ and returns an element $y\in Y$ can be expressed in two stages. Instead of $f(z,x) = y$ we first consider $\lambda(f)\colon Z\rightarrow Y^X$ with $\lambda(f)(z) = f(z,\1)$ which yields a function from $X$ to $Y$ for every $z\in Z$. This first stage is also called currying. It takes a function with two arguments and creates a function of the second argument with the first argument fixed. In the second stage, we then use the evaluation morphism $\varepsilon_{X,Y}$ to apply this function in $Y^X$ to an element in $X$. All these curried functions are also added to $Y^X$.

For any two objects $X,Y \in \Ob(\C)$ in a domain category $\C$ we can add the exponential object. We define $Y^X = \Hom_{\C}(X,Y)\cup \{f(z,\1)\,|\, f\colon Z\times X \rightarrow Y, z\in Z$ for $Z\times X\in \Ob(\C)\}$ and $\varepsilon_{X,Y}(f,x) = f(x)$.
For any $Z\in \Ob(\C)$ we add exactly one morphism $\lambda(f)\colon Z\rightarrow Y^X$ for every $f\colon Z\times X \rightarrow Y$, and define $\varepsilon_{X,Y}\circ(\lambda(f)\times id_X) = f$. Hence, there is a unique $\lambda(f)$ for every $f\colon Z\rightarrow Y^X$ with $\lambda(f)(z) = f(z,\1)$ for every $z\in Z$.
Each domain category contains $1$, $B$ and $E$. More products and exponential objects can be added depending on the domain, for example $E^n$ for n-ary relations  or $B^{E\times E}$ to use objectified binary relations in more complex constructions or add higher-order relations to the domain category as shown in the next subsection. We use the definition of exponentials to add the objects $B^E$ and $B^{E\times E}$ and the corresponding evaluation morphisms $\varepsilon_{E,B}$ and $\varepsilon_{E\times E , B}$ to our example domain category. $B^{E}$ and $B^{E\times E}$ will be used frequently, hence, we will often write $\BE$ and $\BEE$ instead, for easier readability. 

\begin{figure}[h]
    \centering
    \includegraphics[width=0.8\textwidth]{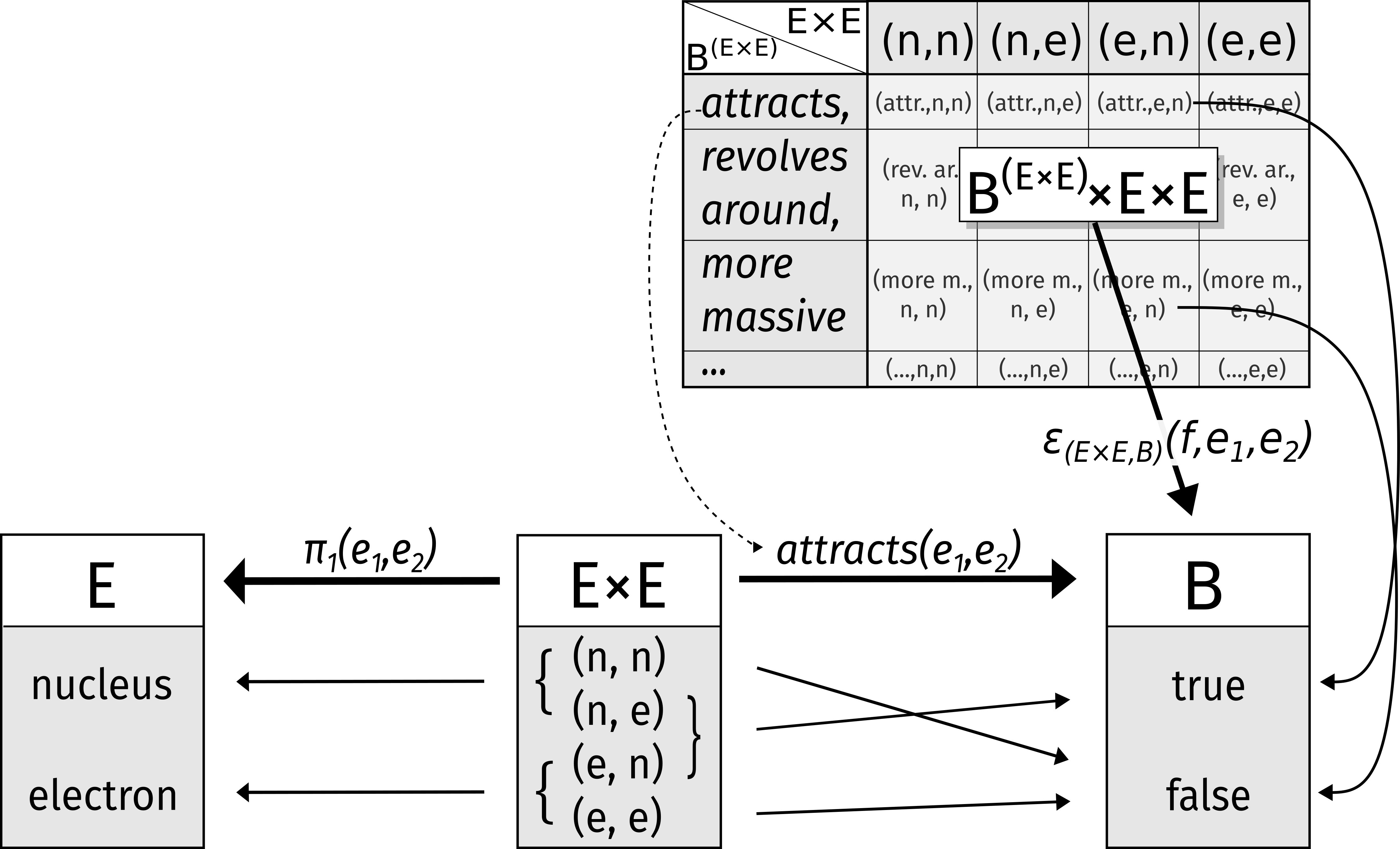}
    \caption{Parts of the atom-category $\atom$ showing $E$ and the product $E\times E$ as well as $B^{(E\times E)}$ and some morphisms.}
    \label{fig:example_BEE}
\end{figure}

\autoref{fig:example_BEE} shows part of the category $\atom$, the domain of the hydrogen atom, as an example. The product $E\times E$ consists of all pairs of the entities $\te{nucleus}$ (n) and $\te{electron}$ (e). The example shows how these are mapped to single entities by the projection to the first component $\pi_1$ and to the elements of $B$ by the binary relation $\te{attracts}(\2)$. The box at the top shows the objects $B^{E\times E}$ containing various binary relations, $E\times E$ and their product that contains tuples, each with a relation and a pair of entities. The evaluation function $\varepsilon_{E\times E, B}$ maps each such tuple to $\te{true}$ or $\te{false}$ in $B$, which is shown for two examples.

\autoref{fig:terminal-exponential} shows a diagram using the projection $\pi_2\colon Z\times X\rightarrow X$ to turn a function $g\colon X\rightarrow Y$ into a function $g' = g\circ \pi_2$. If each morphism from $X$ to $Y$ is represented in $Y^X$ we can define $\lambda(g)(z)$ as the unique morphism mapping any $z$ to $g\in Y^X$ with $\varepsilon_{X,Y}\circ(\lambda (g)\times id_X) = g'$ for any $Z\times X\in \Ob(\C)$.

We can now define a category that describes a domain.

\begin{definition}[Domain Category]
    A domain category $\C$ is a category with sets as objects and named functions on those sets as morphisms.
    Every domain category $\C$ has the objects $B = \{\te{true}, \te{false}\}$, a unit object $1_\C = \{*\}$ and a set of entities $E$. Additionally, there are identity morphisms for every object, morphisms $x\colon 1\rightarrow X$ for every $x\in X$ and any $X\in \Ob(\C)$ and a morphism $f\colon E\rightarrow B$ for every attribute in the domain mapping every entity to $\te{true}$ or $\te{false}$ in $B$.

    A domain category may additionally contain finite Cartesian products and exponential objects consisting of any combination of the three base objects. These are constructed as described above.
    
    To adhere to the properties of a category,  each object is endowed with an identity morphism and all compositions of morphisms $f \colon X\rightarrow Y$ and $g \colon Y\rightarrow Z$ for any $X,Y,Z\in \Ob(\C)$ are added as ${g\circ f}\colon X\rightarrow Z$  if they do not already exist in the category, for example if $X = 1_\C$. 

\end{definition}

We now have defined and introduced the basic building blocks of a domain category, as shown in \autoref{tab:domain_cat}.  For better readability identities and compositions are not included.

\begin{table}[h]
\centering
\begin{tabularx}{1\textwidth}{lXlX}
\toprule
\textit{objects} &  & \multicolumn{2}{l}{\textit{selected morphisms}}  \\
\midrule
$1_\C: $ & $\{*\}$ & $1_\C\rightarrow X$ & $x(*) = x$ for all $x\in X$\\
\hline\\[-1em]
$B:$ & $\{\te{true}$, $\te{false}\}$ & &\\
\hline\\[-1em]
$E:$ & set of entities & $E\rightarrow B$ & attributes, $f\circ \delta $ for $f\colon E\times E\rightarrow B$\\
\hline
\hline\\[-1em]
\makecell[l]{$X\times Y$ \\e.g. $E\times E$\\$= E^2:$} & \makecell[l]{set of all pairs $(x,y)$} & \makecell[l]{$X\times Y\rightarrow X$, $Y$ \\ $Z\rightarrow X\times Y$\\$E^2\rightarrow B$ \\ $E^2\rightarrow E^2$} & \makecell[l]{$\pi_1, \pi_2$\\ $\langle f, g\rangle$\\ all binary  relations\\ $\sigma_{E,E} = \langle\pi_2, \pi_1\rangle$} \\
\hline\\[-1em]
$Y^X:$ & \makecell[l]{$\Hom(X,Y)$\\plus curried morphisms} & \makecell[l]{$Y^X\times X \rightarrow Y$\\ $Z\rightarrow Y^X$} & \makecell[l]{evaluation $\varepsilon_{X,Y}$\\$\lambda(f)(z) = f(z,\1)$ }\\
\bottomrule
\end{tabularx}
\caption{Common objects of every domain category together with some morphisms (above the double line) and optional additional constructions that can be added to better describe a domain.}
\label{tab:domain_cat}
\end{table}

These building blocks resemble those of a topos, which is a category with finite limits and colimits, exponentials and a subobject classifier. The domain categories defined here, however, do only have products and exponential objects necessary for the specific meanings portrayed, not all binary ones. The unit object is also not a terminal object as the unique morphisms to it are not included. It might be possible to extend the construction of a domain category to be a topos, to benefit from their known features, but we opted to keep the categories as small as possible, so they have a structure that will be best matched to similar domain categories in possible analogies.

\subsection{Higher-order Relations}

We will now take a closer look at two higher-order relations, \emph{causes} and \emph{and}, which make objectifications necessary. These higher-order relations can be added as morphisms and elements of objects to a domain category in order to better describe the domain. For example, in the solar system domain, it is important that the attraction between the sun and planets causes planets to revolve around the sun.  We are interested in the types of these relations and the role they play in a specific domain.

Let us first look at \emph{and}. In our formalisation it is a concatenation of two binary relations. Other than in natural language, each relation has to have a specified type in a domain category. The word `and' does not specify what is combined, but a relation \emph{and}() needs a specific type. We will look at two different versions of \emph{and}() here to showcase this. In the domain category $\sun$ of our example, $\te{and}_1$ is a relation that maps two binary relations $f, g\in B^{E\times E}$ to a new binary relation $(\te{f and}_1\te{ g})\in B^{E\times E}$. This $(\te{f and}_1\te{ g})$ is a new binary relation, that has to be specified and added to the set of morphisms and also to $\BEE$. Returning a binary relation instead of one that takes four entities as input means, that the same pair of entities are used for both relations. Using exponential objects, we get $\te{and}_1\colon \BEE\times \BEE\rightarrow \BEE$. As a contrast, another relation $\te{and}_2$ maps two binary relations  $f, g\in B^{E\times E}$ to a new relation $(\te{f and}_2\te{ g})\in B^{E^4}$ that takes four entities as input and returns a boolean value. The type of such higher-order relations can vary for different domain categories and has to be chosen when they are added. 

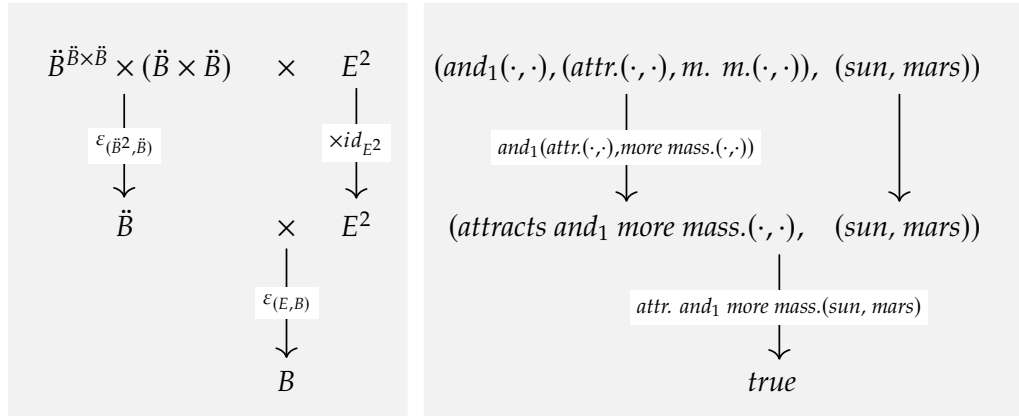
\begin{figure}[h]
    \centering
    \tikz[
overlay]{
    \filldraw[fill=gray!10,draw=black!0] (-6.7,0) rectangle (-1.4,-5.5);
    \filldraw[fill=gray!10,draw=black!0] (-1.2,-0) rectangle (6.8,-5.5);
}
\[\begin{tikzcd}[ampersand replacement=\&,column sep=tiny]
	{{\BEE}^{\BEE\times\BEE}\times (\BEE\times\BEE)} \& {\times } \& {E^2} \&\& {(\te{and}_1(\2), (\te{attr.}(\2), \te{m. m.}(\2)),} \& {\hspace{-1em}(\te{sun, mars}))} \\
	\\
	{\hspace{-1em}\BEE} \& {\times } \& {E^2} \&\& {(\te{attracts and}_1\te{ more mass.}(\2),} \& {\hspace{-1em}(\te{sun, mars}))} \\
	\\
	\& B \&\&\& {\hspace{10em}\te{true}}
	\arrow["{\varepsilon_{({\BEE}^2, \BEE)}}"{description}, shift left=-2,  from=1-1, to=3-1]
	\arrow["{\times id_{E^2}}"{description}, from=1-3, to=3-3]
	\arrow["{\te{and}_1(\te{attr.}(\2),\te{more mass.}(\2))}"{description}, from=1-5, to=3-5]
	\arrow[ shift right=3, from=1-6, to=3-6]
	\arrow["{\varepsilon_{(E²,B)}}"{description}, from=3-2, to=5-2]
	\arrow["{\te{attr. and}_1 \te{ more mass.} (\te{sun, mars})}"{description}, shift left=20, from=3-5, to=5-5]
\end{tikzcd}\]
    \caption{$\te{and}_1$ as object: the left side shows the progression from the complex type via two evaluations to a Boolean, the right side shows how an example tuple is evaluated.}
    \label{fig:and_example}
\end{figure}

In order to keep the set of morphisms from $E\times E$ to $B$ or from $E^4$ to $B$ finite, we can decide that \emph{and} is commutative and associative, so every combination of relations occurs only once in $\BEE$. We can now add a new object to $\sun$ containing the objectified version of $\te{and}_1$, namely $\BEE^{\BEE\times \BEE}$. \autoref{fig:and_example} shows how the product $\BEE^{\BEE\times \BEE}\times \BEE\times \BEE\times E^2$, that is also part of the domain category $\sun$, is evaluated in two steps and what that means for one example tuple.

Now let us add the morphism $\te{causes}(\2)$, which has the same type as $\te{and}_1(\2)$. We will use \emph{causes(attracts (sun, mars), revolves around(mars, sun))} as an example. If possible it is desirable to bring the expression into a form with the same entities added, so here $\te{causes(attracts}(e_1,e_2)\te{, (revolves around} \circ\sigma_{E\times E})(e_1,e_2))$ which takes a pair of entities and returns a boolean value, so it has type $B^{E\times E}$. This form constricts the set of possible combinations of cause and effect to ones where the entities involved in both are the same. Whether the set of possible combinations is constricted to the same two entities as input is a design choice like it was with \emph{and} above. In this case, it is a reasonable assumption that cause and effect are related to the same entities and it is a nice feature of the formalism that this assumption can be captured in this way.

\begin{figure}[h]
    \centering
    \includegraphics[width=0.8\textwidth]{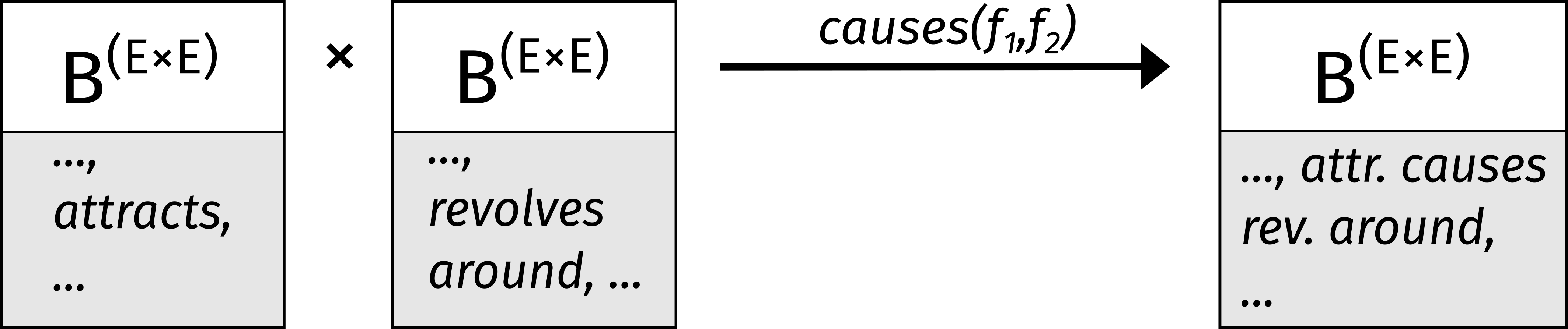}
    \caption{The morphism \textit{causes} within a category.}
    \label{fig:example_causes}
\end{figure}

\begin{figure}[h]
    \centering
    \tikz[
overlay]{
    \filldraw[fill=gray!10,draw=black!0] (-6.9,0) rectangle (-2.4,-5.5);
    \filldraw[fill=gray!10,draw=black!0] (-2.2,-0) rectangle (7.0,-5.5);
}
\[\begin{tikzcd}[ampersand replacement=\&,column sep=tiny]
	{{\BEE}^{\BEE^2}\times (\BEE^2)} \& {\times } \& {E^2} \&\& {(\te{caus.}(\2),(\te{attr.}(\2),\te{rev. ar.}\circ \sigma_{E,E}(\2)),} \& {\hspace{-1em}\te{(sun, mars))}} \\
	\\
	{\BEE} \& {\times } \& {E^2} \&\& {(\te{attr. causes rev. ar.}\circ\sigma_{E\times E}(\2),} \& {\hspace{-1em}\te{(sun, mars))}} \\
	\\
	\& B \&\&\& {\hspace{10em}\te{true}}
	\arrow["{\varepsilon_{({\BEE}^2, \BEE)}}"{description},  from=1-1, to=3-1]
	\arrow["{\times id_{E^2}}"{description}, from=1-3, to=3-3]
	\arrow["{\te{causes}(\te{attr.}(\2),\te{rev. ar.}\circ \sigma_{E,E}(\2))}"{description}, from=1-5, to=3-5]
	\arrow[ shift right=3, from=1-6, to=3-6]
	\arrow["{\varepsilon_{(E²,B)}}"{description}, from=3-2, to=5-2]
	\arrow["{\te{attr. causes rev. ar.}\circ\sigma_{E\times E}(\te{sun, mars})}"{description}, shift left=20, from=3-5, to=5-5]
\end{tikzcd}\]
    \caption{$\te{causes}$ as object: the left side shows the progression from the complex type via two evaluations to a Boolean, the right side shows the elements in these objects which represent the sentence `the sun attracts mars, which causes mars to revolve around the sun'.}
    \label{fig:causes}        
  
\end{figure}
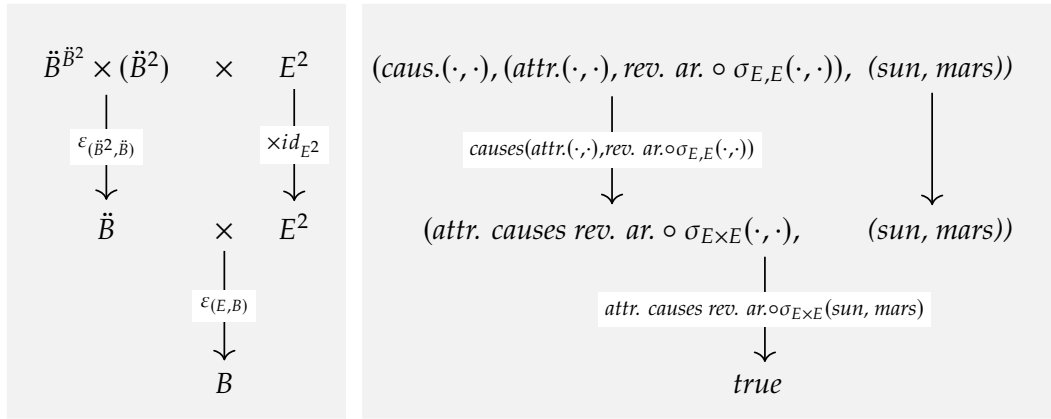

The types of the relations that are involved are $B^{E\times E}$ for \emph{attracts} and \emph{revolves around} $\circ\sigma_{E\times E}$.
We can combine these observations of types of the different relations and think about the type of this particular \emph{causes}. We see that, like $\te{and}_1$, it takes two morphisms $f,g\in\BEE$  and returns a new morphism $f\te{ causes }g\in \BEE$ as shown in \autoref{fig:example_causes}. These new morphisms are added to the category and are also elements in $\BEE$. \autoref{fig:causes} shows the involved objects and their evaluation to a Boolean value. 

\subsection{Categories for the solar system and hydrogen atom}
We have now described all building blocks needed to construct a domain category. In this section we will recap these and construct the categories describing the solar system and the hydrogen atom. 
Remember that each domain category $\C$ consists of an object of entities $E$, a boolean object $B$ and a unit object $1_\C$. These objects come equipped with identity morphisms, as well as some basic morphisms listed in \autoref{tab:domain_cat}.

The domain $\sun$ describing the solar system contains the entities $E_\sun=\{\te{sun}, \te{mars}, \te{venus}\}$ and the domain $\atom$ contains the entities $E_\atom =\{\te{nucleus}, \te{electron}\}$. All the objects are listed in \autoref{tab:solar_cat}.

Attributes are represented in a domain category as morphisms from $E$ to $B$, while relations with more entities are included using products of the entity object with itself. The domains of the solar system and the atom include binary relations, therefore the binary product $E\times E$ is added to the categories $\sun$ and $\atom$. For these two domains we do not need relations with an arity greater than two, hence $E^n$ for $n$ greater $2$ is not added to the categories.

Higher-order relations are included using exponential objects, which are sets of objectified morphisms and enable us to describe relations on relations. For each domain, only the exponential objects that are needed to describe the higher-order relations that occur in that domain are added. Additionally, all morphisms that are necessary to fulfil the properties of a category, products and exponentials are added, that is identities, compositions, projections, evaluations and the unique morphisms fulfilling the properties of products and exponential objects.
\autoref{tab:solar_cat} lists al objects of the categories $\sun$ and $\atom$ and for each object the elements in that object.

\begin{figure}
    \centering
\[\begin{tikzcd}[ampersand replacement=\&]
	\& {B^E} \& {B^{E}\times E} \\
	\&\& E \\
	1 \&\&\& B \\
	\&\& {E\times E} \\
	{\BEE^2} \& \BEE \& {B^{E^2}\times E^2} \\
	{\BEE^{\BEE^2}} \& {\BEE^{\BEE^2}\times\BEE^2} \& {(\BEE^{\BEE^2}\times\BEE^2)\times E^2}
	\arrow[from=1-3, to=1-2]
	\arrow[shift right, from=1-3, to=2-3]
	\arrow[from=1-3, to=3-4]
	\arrow[from=2-3, to=1-2]
	\arrow[shift right, from=2-3, to=1-3]
	\arrow[from=2-3, to=3-4]
	\arrow[shift right, from=2-3, to=4-3]
	\arrow[from=3-1, to=1-2]
	\arrow[from=3-1, to=1-3]
	\arrow[from=3-1, to=2-3]
	\arrow[from=3-1, to=3-4]
	\arrow[from=3-1, to=4-3]
	\arrow[from=3-1, to=5-1]
	\arrow[from=3-1, to=5-2]
	\arrow[from=3-1, to=5-3]
	\arrow[curve={height=12pt}, from=3-1, to=6-1]
	\arrow[from=3-1, to=6-2]
	\arrow[from=3-1, to=6-3]
	\arrow[shift right, curve={height=12pt}, from=4-3, to=1-3]
	\arrow[shift right, from=4-3, to=2-3]
	\arrow[from=4-3, to=3-4]
	\arrow[from=5-1, to=5-2]
	\arrow[from=5-3, to=3-4]
	\arrow[from=5-3, to=4-3]
	\arrow[from=5-3, to=5-2]
	\arrow[from=6-2, to=5-1]
	\arrow[from=6-2, to=5-2]
	\arrow[from=6-2, to=6-1]
	\arrow[curve={height=24pt}, from=6-3, to=4-3]
	\arrow[from=6-3, to=5-3]
	\arrow[from=6-3, to=6-2]
\end{tikzcd}\]

    \caption{Objects and morphisms in the domain category $\sun$: The category includes $\BEE^{\BEE^2}\times\BEE^2\times E^2$ as used in \autoref{fig:causes} and  the parts it is made up of. There is an arrow between two objects if there is at least one morphism between them. This does not include compositions, identities and other morphism from an object to itself.}
    \label{fig:domain-cat-ex}
\end{figure}
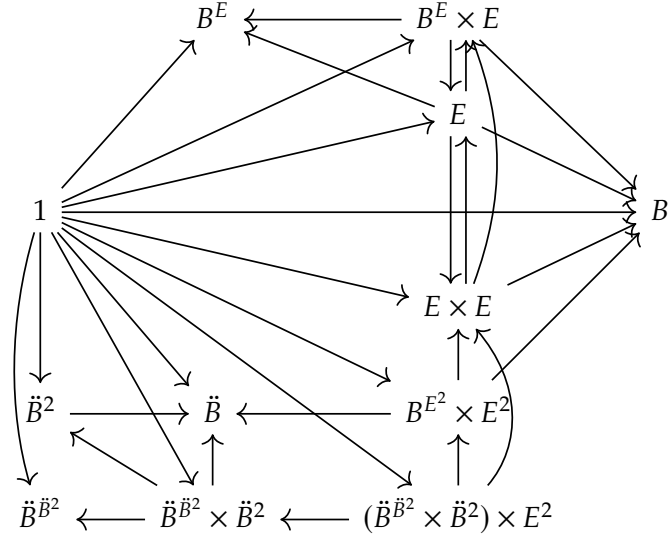

All in all, we have defined a basic structure for domain categories that can capture n-ary relations and higher-order predicates, but this structure still leaves many design choices for each individual domain. Such choices include specific higher-order relations like \emph{and} and \emph{causes} and their types as well as the morphisms they map to. \autoref{fig:domain-cat-ex} shows the structure of $\sun$ with all products and exponentials added to model \emph{causes} and \emph{and}$_1$. Other domains can include other complex objects like $E^3$ or $B^{E^4}$, that would be added similarly.

\begin{table}[h]
\begin{tabularx}{\textwidth}{lXX}
\toprule
\textit{object} & \textit{elements in $\sun$} & \textit{elements in $\atom$}\\
\midrule 
$1$ & $*$ & $*$\\
\hline\\[-1em]
$B$ & \emph{true}, \emph{false} & \emph{true}, \emph{false}\\
\hline\\[-1em]
 $E$ & \makecell[l]{\emph{sun, mars, venus}} & \makecell[l]{\emph{nucleus, electron}}\\
\hline\\[-1em]
$E\times E = E^2$ & \makecell[l]{\emph{(sun, sun)}, \emph{(sun, mars)}, \\\emph{(sun, venus)}, \emph{(mars, sun)},\\ \emph{(mars, mars)}, \emph{(mars, venus)}\\
 \emph{(venus, sun)}, \emph{(veuns, mars)}, \\\emph{(venus, venus)}} & \makecell[l]{\emph{(nucleus, nucleus)},\\ \emph{(nucleus, electron)}, \\\emph{(electron, nucleus)},\\\emph{(electron, electron)}}\\
 \hline\\[-1em]
$B^E = \BE$ & \makecell[l]{\emph{hot}, $f\circ\delta$ for $f\in \BEE$\\ $f(e,\1)$ for $e\in E$ and $f\in \BEE$}& \makecell[l]{$f\circ\delta$ for $f\in \BEE$\\ $f(e,\1)$ for $e\in E$ and $f\in \BEE$}\\
\hline\\[-1em]
$B^E\times E$ & $(f,e)$ for all $f\in \BE$ and $e\in E$ & $(f,e)$ for all $f\in\BE$ and $e\in E$ \\
\hline\\[-1em]
$B^{E\times E} = \BEE$& \makecell[l]{\emph{attracts}, \emph{more massive},\\\emph{revolves around}, $f\circ\sigma$,\\ $(f \te{ causes }g)$, $(f\te{ and }g)$\\ for $f,g \in \BEE$,  $f\circ \pi_i$\\ for $i\in\{1,2\}$ and $f\in \BE$} 
& \makecell[l]{\emph{attracts}, \emph{more massive},\\\emph{revolves around}, $f\circ\sigma$,\\ $(f\te{ causes }g)$, $(f\te{ and }g)$\\ for $f,g \in \BEE$,  $f\circ \pi_i$\\ for $i\in\{1,2\}$ and $f\in \BE$}\\
\hline\\[-1em]
$\BEE^{\BEE\times \BEE}$ & \makecell[l]{\emph{causes}, \emph{and}, $\pi_1$, $\pi_2$}  &\makecell[l]{\emph{causes}, \emph{and}, $\pi_1$, $\pi_2$} \\
\hline\\[-1em]
\makecell[l]{$\BEE\times \BEE$, $\BEE\times E^2$\\ $\BEE^{\BEE^2}\times\BEE^2$\\  $(\BEE^{\BEE^2}\times\BEE^2)\times E^2$}  & \makecell[l]{respective tuples e.g. $B^{E^2}\times E^2$\\$=\{(\te{more mass.}(\2),(\te{mars, sun}))$,\\$(\te{attracts}(\2),(\te{sun, mars})),\dots\}$}  & \makecell[l]{all respective tuples} \\
\bottomrule
\end{tabularx}
\caption{Objects in the categories for the solar system and atom.}
\label{tab:solar_cat}
\end{table}

\section{Analogies between Domain Categories}\label{sec:analogy}
So far, we have discussed how types can be used to make the different building blocks of a domain more explicit and to form higher-order relations. We have constructed the two domains for an analogy, but have not compared the two domains to each other, yet.

\subsection{Functor: Comparing Domains}
This section deals with comparing and matching the two domain categories that model the base and target of an analogy. Part of making an analogy is finding similarities between the domains to guide a possible transfer.
In \ac{ct}, the simplest way to compare two domain categories is to look at functors between them. These are structure preserving maps between categories.

\begin{definition}[Functor]
    A functor $F\colon  \C \rightarrow \D$ consists of an object part $F_{\Ob}$ that maps all objects in $\Ob(\C)$ to ones in $\Ob(\D)$ and a family of functions that map the morphisms. For each pair of objects $X,Y\in \Ob(\C)$ there is a function $F_{XY}\colon \Hom_{\C}(X,Y)\rightarrow \Hom_{\D}(F(X),F(Y))$ that maps the morphisms from $X$ to $Y$ to morphisms between their images $F(X)$ and $F(Y)$.

    Additionally a functor  is:
    \begin{itemize}
        \item identity preserving: $F_{XX}(id_X) = id_{F(X)}$ for all $X\in\Ob(\C)$
        \item composition preserving: $F_{XZ}(f\circ g) = F_{YZ}(f)\circ F_{XY}(g)$ for all morphisms $f,g$ in $\C$.
    \end{itemize}
\end{definition}

We omit the indices when they are not needed for better readability.
As mentioned above, all domain categories are small categories because they have a set of objects.
The category of small categories $\cat$ has the class of all small categories as objects and functors as morphisms. It is easy to verify, that $\cat$ fulfills all properties of a category. The domain categories constructed here are also objects in $\cat$.

The most trivial functor maps all objects of one category to one single object $X$ in the target domain and all morphisms to its identity $id_X$. This functor is identity preserving because $F(id_Y) = id_X = id_{F(Y)}$ for any object $Y$ in the domain category. It is also composition preserving because $F(f\circ g) = id_X = id_X\circ id_X = F(f)\circ F(g)$ for all morphisms $f$ and $g$ in the domain category.
Sometimes it is not possible to define a more interesting functor between two categories, hence, in these cases we will use partial functors.

\begin{definition}[Partial functor]
A partial functor $F|_{C'}\colon \C \rightarrow \D$ is a functor from a subcategory (see Definition \ref{def:subcategory}) $\C'$ of $\C$ to $\D$.    
\end{definition}

There can be many different functors between two categories, but not all of them are suited to describe an analogy. We will now add different properties that a functor must have in order to be considered an analogy.

The first property is preserving truth. If an attribute is true for a specific entity, then the image of this attribute should also be true for the image of that entity. In each domain category, there are the objects $1$ and $B$ and morphisms $\te{true}\colon 1\rightarrow B$ and $\te{false}\colon 1\rightarrow B$. A functor $F$ from a category $\C$ to another category $\D$ describing an analogy has to map the morphism $\te{true}_C$ to $\te{true}_D$ and $\te{false}_C$ to $\te{false}_D$. We call this property \emph{truth preserving}. It follows that $F(1_\C) = 1_\D$ and $F(B_\C) = B_\D$.

As an example, we will now construct an analogy between the domain categories $\sun$ and $\atom$ that describe the solar system and the hydrogen atom, respectively.

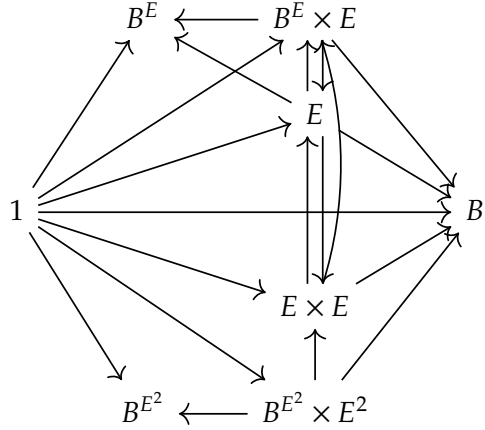
\begin{figure}[h]
    \centering
\[\begin{tikzcd}[ampersand replacement=\&]
	\& {B^E} \& {B^{E}\times E} \\
	\&\& E \\
	1 \&\&\& B \\
	\&\& {E\times E} \\
	\& {B^{E^2}} \& {B^{E^2}\times E^2}
	\arrow[from=1-3, to=1-2]
	\arrow[shift left, from=1-3, to=2-3]
	\arrow[from=1-3, to=3-4]
	\arrow[from=2-3, to=1-2]
	\arrow[shift left, from=2-3, to=1-3]
	\arrow[from=2-3, to=3-4]
	\arrow[shift left, from=2-3, to=4-3]
	\arrow[from=3-1, to=1-2]
	\arrow[from=3-1, to=1-3]
	\arrow[from=3-1, to=2-3]
	\arrow[from=3-1, to=3-4]
	\arrow[from=3-1, to=4-3]
	\arrow[from=3-1, to=5-2]
	\arrow[from=3-1, to=5-3]
	\arrow[curve={height=12pt}, from=4-3, to=1-3]
	\arrow[shift left, from=4-3, to=2-3]
	\arrow[from=4-3, to=3-4]
	\arrow[from=5-3, to=3-4]
	\arrow[from=5-3, to=4-3]
	\arrow[from=5-3, to=5-2]
\end{tikzcd}\]
    \caption{The basic structure of a domain category with some additional products and exponential objects. There is an arrow from $X$ to $Y$ if there is at least one morphism $g\colon X\rightarrow Y$. Identities and compositions were omitted for better readability. }
    \label{fig:simple-cat}
\end{figure}

\begin{figure}
    \centering
\[\begin{tikzcd}[ampersand replacement=\&]
	\&\& {E_\sun} \&\&\&\& {E_\atom\times E_\atom = F(E_\sun)} \\
	\\
	{1_\sun} \&\& {E_\sun\times E_\sun} \&\& {1_\atom} \&\& {E_\atom = F(E_\sun\times E_\sun)} \\
	\\
	\&\& {E_\sun} \&\&\&\& {E_\atom\times E_\atom = F(E_\sun)}
	\arrow["{s_1}", from=3-1, to=1-3]
	\arrow["{(s_1,s_2)}", from=3-1, to=3-3]
	\arrow["{s_2}"', from=3-1, to=5-3]
	\arrow["{\pi_1}"', from=3-3, to=1-3]
	\arrow["\pi2", from=3-3, to=5-3]
	\arrow["{(a,a) = \delta\circ a= F(s_1)}"{description}, from=3-5, to=1-7]
	\arrow["{a =}", from=3-5, to=3-7]
	\arrow["{ F(s_1,s_2)}"', draw=none, from=3-5, to=3-7]
	\arrow["{(a,a) = \delta\circ a = F(s_2)}"{description}, from=3-5, to=5-7]
	\arrow["\delta"', from=3-7, to=1-7]
	\arrow["\delta", from=3-7, to=5-7]
\end{tikzcd}\]
    \caption{Possible functor $F\colon \sun\rightarrow \atom$ that maps $1_\sun$ to $1_\atom$, but $E_\sun$ to $E_\atom\times E_\atom$ and $E_\sun\times E_\sun$ to $E_\atom$. The functor hast to preserve compositions ($F(f\circ g) = F(f)\circ F(g)$), so morphisms describing entities in $\sun$ can only be mapped to pairs of the same entity in $\atom$.}
    \label{fig:functor-ex}
\end{figure}
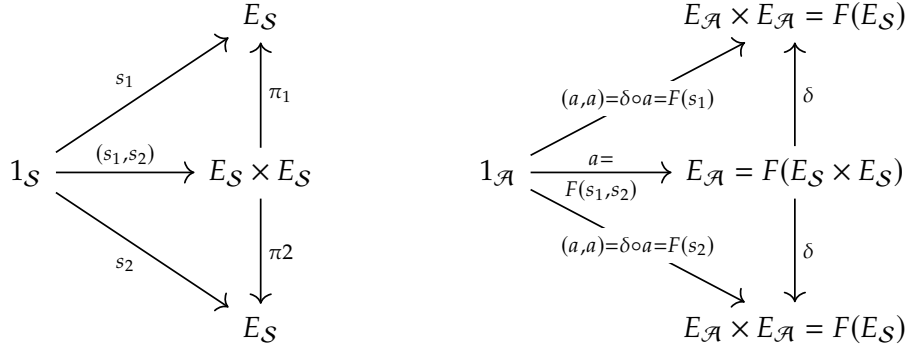

We construct the functor $F\colon \sun \rightarrow \atom$ with an injective function as the object part $F_{\Ob}$ and see which object mappings follow from the structure of the two categories. 
\autoref{fig:simple-cat} shows the basic domain category in a simplified way. There is an arrow between two objects if there exists at least one morphism between them. A functor $F$ can map two objects $X$ and $Y$ with a non-empty morphism set $\Hom(X,Y)$ only to a pair of objects $F(X)$ and $ F(Y)$ which also have at least one morphism $f\colon F(X)\rightarrow F(Y)$. 
As mentioned above, truth preserving ensures that $1_\sun$ is mapped to $1_\atom$ and that $B_\sun$ is mapped to $B_\atom$. We have to take a closer look at the morphisms for the maps of $E$ and $E\times E$. 
We have noted that the object mappings are not functions. However, both domain categories contain a unit object $1$ with morphisms from this object to $E$ for each element of $E$. So in our example there are morphisms $\te{sun}(\1)$, $\te{mars}(\1)$ and $\te{venus}(\1)$ from $1_\sun$ to $E_\sun$ and $\te{nucleus}(\1)$ and $\te{electron}(\1)$ from $1_\atom$ to $E_\atom$, that have to be mapped to each other by a functor that maps $1_\sun$ to $1_\atom$ and $E_\sun$ to $E_\atom$.
In theory it might be possible to map $E_\sun$ to $E_\atom\times E_\atom$ and $E_\sun\times E_\sun$ to $E_\atom$. In this case, the morphisms $\te{sun}$, $\te{mars}$ and $\te{venus}\in\Hom(1, E)$ would have to be mapped to a morphism from $1_\atom$ to $E_\atom\times E_\atom$ such as $(\te{nucleus, nucleus})(\1)$ or $\te{(electron, nucleus)}(\1)$. Similarly $\pi_1$ and $\pi_2$ would have to be mapped to $\delta$, the only morphism from $E$ to $E\times E$. \autoref{fig:functor-ex} shows that the only way to have such a functor that is composition preserving, is to map all pairs of \emph{sun} and planets to the same pair of either \emph{nucleus} or \emph{electron}. However, this will lead to problems with the relation morphisms like $\te{attracts}\colon E\times E\rightarrow B$.

This is just one example domain and we cannot be certain that there will always be a unique object function.
It is not a general property of functors that they necessarily map products to products and exponential objects to exponential objects, but we add this consistency property to our definition of a functor describing an analogy. So it applies for any functor $F\colon \C\rightarrow \D$ and $X,Y\in \Ob(\C)$ that $F(X\times Y) = F(X)\times F(Y)$ with $F(\pi_1) = \pi_1$ and $F(Y^X)= F(Y)^{F(X)}$ with $F(\varepsilon_{X,Y})=\varepsilon_{F(X),F(Y)}$.

All in all we get the following definition of an analogy.

\begin{definition}[Analogy]\label{def:analogy}
    A $\C$-$\D$ analogy from domain category $\C$ describing the base to domain category $\D$ describing the target is a (partial) functor $F\colon \C\rightarrow \D$ with injective object part and the following properties:
\begin{itemize}
    \item truth preserving: $F(true_\C(\1)) = true_\D(\1)$ and $F(false_\C(\1)) = false_\D(\1)$
    \item product preserving: $F(X\times Y) = F(X) \times F(Y)$ for any $X,Y\in \Ob(\C)$
    \item exponential preserving: $F(Y^X) = F(Y)^{F(X)}$ and $F(\varepsilon_{X,Y}) = \varepsilon_{F(X),F(Y)}$ 
\end{itemize}
\end{definition}

As a functor, $F$ is also identity preserving and composition preserving.
A functor $F\colon \sun\rightarrow\atom$ that fulfills these constraints might map all objects according to the rows in \autoref{tab:solar_cat} that lists all the objects in the categories $\sun$ and $\atom$. 

It is important to note that the single object mappings are not functions between sets. Saying that $F(E_\sun)$ equals $E_\atom$ states that the sets of entities correspond, but not that  single entities are mapped to each other. 

\begin{figure}[h]
    \centering
\[\begin{tikzcd}[ampersand replacement=\&]
	{Y^X\times X} \&\& Y \&\& {F(Y)^{F(X)}\times F(X)} \&\& {F(Y)} \\
	\\
	{Z\times X} \&\&\&\& {F(Z)\times F(X)}
	\arrow[""{name=0, anchor=center, inner sep=0}, "{\varepsilon_{X,Y}}", from=1-1, to=1-3]
	\arrow["F"{description}, curve={height=-30pt}, dotted, from=1-1, to=1-5]
	\arrow[""{name=1, anchor=center, inner sep=0}, "{\varepsilon_{F(X),F(Y)}}", from=1-5, to=1-7]
	\arrow[""{name=2, anchor=center, inner sep=0}, "{\lambda (f)\times id_X}", from=3-1, to=1-1]
	\arrow["f"', from=3-1, to=1-3]
	\arrow["F"{description}, curve={height=18pt}, dotted, from=3-1, to=3-5]
	\arrow[""{name=3, anchor=center, inner sep=0}, "{\lambda(F(f))\times id_{F(X)}}", from=3-5, to=1-5]
	\arrow["{F(f)}"', from=3-5, to=1-7]
	\arrow["F"{description}, curve={height=18pt}, shorten <=18pt, shorten >=18pt, dotted, from=0, to=1]
	\arrow["F"{description, pos=0.4}, shorten <=16pt, shorten >=63pt, dotted, from=2, to=3]
\end{tikzcd}\]
    \caption{A functor $F$ mapping an exponential object, evaluation and corresponding maps in one domain to an exponential object, evaluation and corresponding maps in another.}
    \label{fig:F-lambda}
\end{figure}
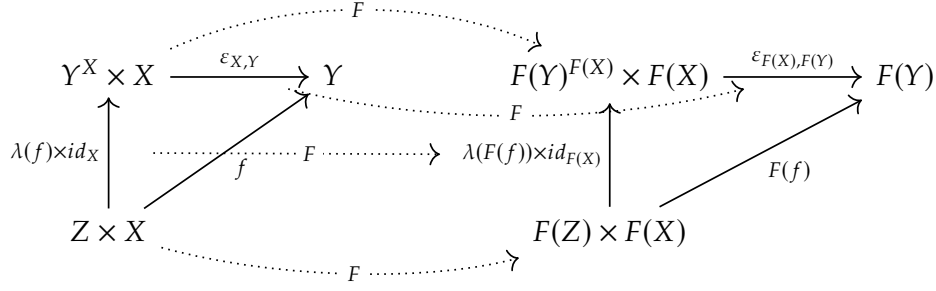

A preservation of composition  and exponential objects implies for any $f\colon Z\times X\rightarrow Y$ with $F(f) = g\colon F(Z) \times F(X) \rightarrow F(Y)$ and $Y^X\in \Ob(\C)$ and $F(Y)^{F(X)}\in \Ob(\D)$, that $F(\lambda(f)) = \lambda(g)$, because $\lambda(g)$ is the unique morphism with $\varepsilon_{F(X),F(Y)}\circ(\lambda(g)\times id_{F(X)}) = g = F(f)$ (see \autoref{fig:F-lambda}). So there is a consistency within morphism maps. However, similar to the object part, these maps are not concerned yet with how the morphisms act. So far, it is possible to map the morphism $\te{attracts}(\2)$ in $\sun$ to $\te{revolves around}(\2)$ in $\atom$, as long as this is done consistently throughout the category and the entity morphisms from $1$ to $E$ are also mapped to preserve compositions.

We have mentioned above that compositions of morphisms $f$ and $g$ are also part of a domain category to adhere to the properties of a category and are usually named $g\circ f$, if there is no other morphism defined as their composition. However, there is only one unique morphism $x\colon 1\rightarrow X$ for every element $x\in X$. So for every $x\colon 1\rightarrow X$ and $f\colon X\rightarrow Y$ with $f(x) = y\in Y$ the rule is that $f\circ x = y \colon 1 \rightarrow Y$. This means that for example $\te{attracts}(\2)\circ(\te{sun}, \te{mars})(\1)=\te{~true}(\1)$ for the morphisms $\te{attracts}(\2)\colon E\times E\rightarrow B$ and $(\te{sun}, \te{mars})(\1)\colon 1\rightarrow E\times E$. In this way morphisms from $1_\sun$ to $E_\sun$ or, as in this example, $E_\sun\times E_\sun$ and morphisms from $E_\sun$ or $E_\sun\times E_\sun$ to $B_\sun$ must be mapped to morphisms in $\atom$ in a consistent way, meaning that if a composition is true the composition of the maps must also be true. 

The columns in \autoref{tab:ex_analogy} represent an example map of morphisms $1_\atom\rightarrow E_\atom\times E_\atom$ representing pairs of entities in $\atom$ to morphisms $1_\sun\rightarrow E_\sun\times E_\sun$ in $\sun$. A map of binary relations in $\atom$ to binary relations in $\sun$ is shown in the rows. The entries in the table show whether the composition of the morphism $1\rightarrow E\times E$ in that column and the binary relation $E\times E \rightarrow B$ is the morphism \emph{true}$\colon 1\rightarrow B$ ($t$) or \emph{false}$\colon 1\rightarrow B$ ($f$). Each entry shows the boolean value for $\atom$ (top row and left column) and $\sun$ (bottom row and right column). We can see that the map suggested here, that maps binary relations to relations with the same name is truth preserving, because in each cell both entries are either $\te{true}(\1)$ or $\te{false}(\1)$.

\begin{table}[h]
\centering
\begin{tabularx}{1\textwidth}{ll|XXXX|ll}
 & & \multicolumn{4}{c|}{$1_\atom \rightarrow E_\atom \times E_\atom$} & & \\
 & & $(n, n)$ & $(n, e)$ & $(e, n)$ & $(e, e)$ & & \\
\hline
\multirow{4}{*}{\begin{sideways} $E^2_{\atom}\rightarrow B_{\atom}$ \end{sideways}}&$\te{attracts}$& $f_\atom$, $f_\sun$ & $t_\atom$, $t_\sun$ & $t_\atom$, $t_\sun$ & $f_\atom$, $f_\sun$ & $\te{attracts}$ & \multirow{4}{*}{\begin{sideways} $E^2_{\sun}\rightarrow B_{\sun}$ \end{sideways}}\\
&$\te{rev. ar.}$& $f_\atom$, $f_\sun$ & $f_\atom$, $f_\sun$ & $t_\atom$, $t_\sun$ & $f_\atom$, $f_\sun$ & $\te{rev. ar.}$ & \\
&$\te{more mas.}$& $f_\atom$, $f_\sun$ & $t_\atom$, $t_\sun$ & $f_\atom$, $f_\sun$ & $f_\atom$, $f_\sun$ & $\te{more mas.}$ & \\
\cdashline{3-6}
&$\te{(attr. causes r. a.)}$& $f_\atom$, $f_\sun$ & $t_\atom$, $t_\sun$ & $f_\atom$, $f_\sun$ & $f_\atom$, $f_\sun$ & $\te{(attr. causes r.a.)}$&  \\
\hline
&&$(s, s)$&\makecell[l]{$(s, v)$,\\$(s, m)$}&\makecell[l]{$(v, s)$,\\$(m, s)$}&\makecell[l]{$(v, v)$,\\$(m, m)$}&&\\
&&\multicolumn{4}{c|}{$1_\sun\rightarrow E_\sun \times E_\sun$}&&\\
\end{tabularx}
\caption{Part of an example analogy: the columns show which morphisms between the unit object and entity product are mapped in an analogy between $\atom$ and $\sun$ while the rows show maps between morphisms representing binary relations. Each cell shows the boolean value of the composition of the morphisms in both domain categories.}
\label{tab:ex_analogy}
\end{table}

There are other possible maps that can be truth preserving. It is for example possible to map $\te{more massive}(\2)$ in $\sun$ to $\te{revolves around}(\2)$ in $\atom$ and vice versa. To make everything consistent one has to map $\te{sun}(\1)$ to $\te{electron}(\1)$ and $\te{mars}(\1)$ and $\te{venus}(\1)$ to $\te{nucleus}(\1)$ as shown in \autoref{tab:ex_analogy_wrong}. We can see, that the entries of boolean values in the first three rows remain the same and consistent. However, this switch has to be followed consistently throughout the category. This leads to inconsistencies in the higher-order relations. The morphism $\te{causes}(\2)$ brings a distinction between revolves around and more massive as elements in $\BEE$ as can be seen below the dashed line in \autoref{tab:ex_analogy_wrong}. The composition $(\te{attracts causes revolves around})\circ(\te{nucleus}, \te{electron})$ is $\te{true}$, while the composition $(\te{attracts causes more massive})\circ(\te{venus},\te{sun})$ is $\te{false}$.

\begin{table}[h]
\centering
\begin{tabularx}{1\textwidth}{ll|XXXX|ll}
 & & \multicolumn{4}{c|}{$1_\atom \rightarrow E_\atom \times E_\atom$} & & \\
 & & $(n, n)$& $(n, e)$ & $(e, n)$ & $(e, e)$ & & \\
\hline
\multirow{4}{*}{\begin{sideways} $E^2_{\atom}\rightarrow B_{\atom}$ \end{sideways}}&$\te{attracts}$& $f_\atom$, $f_\sun$ & $t_\atom$, $t_\sun$ & $t_\atom$, $t_\sun$ & $f_\atom$, $f_\sun$ & $\te{attracts}$ & \multirow{4}{*}{\begin{sideways} $E^2_{\sun}\rightarrow B_{\sun}$ \end{sideways}}\\
&$\te{rev. ar.}$& $f_\atom$, $f_\sun$ & $f_\atom$, $f_\sun$ & $t_\atom$, $t_\sun$ & $f_\atom$, $f_\sun$ & $\te{more mas.}$ & \\
&$\te{more mas.}$& $f_\atom$, $f_\sun$ & $t_\atom$, $t_\sun$ & $f_\atom$, $f_\sun$ & $f_\atom$, $f_\sun$ & $\te{rev. ar.}$ & \\
\cdashline{3-6}
&$(\te{attr. causes r. a.})$& $f_\atom$, $f_\sun$ & $t_\atom$, $f_\sun$ & $f_\atom$, $f_\sun$ & $f_\atom$, $f_\sun$ & $(\te{attr. causes m.m.})$&  \\
\hline
&&\makecell[l]{$(v, v)$,\\$(m, m)$}&\makecell[l]{$(v, s)$,\\$(m, s)$}&\makecell[l]{$(s, v)$,\\$(s, m)$}&$(s, s)$&&\\
&&\multicolumn{4}{c|}{$1_\sun\rightarrow E_\sun \times E_\sun$}&&\\
\end{tabularx}
\caption{Part of an alternative example analogy similar to \autoref{tab:ex_analogy}: morphisms in $\sun$ were changed such that \emph{more massive} is mapped to \emph{revolves around} and vice versa and the entities were permuted accordingly. This remains truth preserving except for \emph{causes}.}
\label{tab:ex_analogy_wrong}
\end{table}

The attribute $\te{hot}(\1)$ in $\sun$ does not have an obvious counterpart in $\atom$. It might be mapped to a concatenation with $\delta$ like $\te{attract}\circ\delta\colon E \rightarrow B$, but this is not consistent, because $\te{hot}\circ \te{sun} = \te{true}$, while $\te{attract}\circ \delta\circ \te{sun} = \te{false}$. As $\te{hot}(\1)$ cannot be mapped into the target domain we do not want it to be a part of the analogy. This can be achieved by using a partial functor that does not include $\te{hot}(\1)$. 

There is a difference between functors from $\sun$ to $\atom$ and those from $\atom$ to $\sun$. These functors are also not unique, even if we require preservation of truth. The morphism $\te{electron}(\1)\colon 1_\atom \rightarrow E_\atom$ can be mapped to $\te{mars}(\1)$ or $\te{venus}(\1)\in\Hom_\sun(1,E)$, but not to both at the same time. So there are at least two equally valid functors from $\atom$ to $\sun$.

Definition \ref{def:analogy} includes partial functors, which will lead to more interesting mappings and are an important part of describing analogies in this way. However, partial functors do not in general adhere to the composition rules for functors. For partial functors to be closed under composition and stable under pullbacks, the inclusion of the subcategory must be a cosieve \citep{benabou_2000}, which is not generally the case for the construction used here.

\citet{diaconescu201732-article} models blending with $\frac{3}{2}$-instituitions based on Gougen's $\frac{3}{2}$-Cate\-gories, whose partial functions propagate to partial or lax functors. \citet{SCHORLEMMER2021118} use the stricter notion of amalgams as categories that has more computational support to define blending. These are two approaches that deal with the partial nature of blends, which could also inspire future directions to deal with the partiality of the analogy functors defined here.

We have now constructed domain categories and used them to define analogies. In the remainder of this section we will use these analogy-functors to construct or find new categories that describe these analogies or the resulting blend of two domains. We will use the two concepts pullback and pushout from category theory.

\subsection{Core and Blend}
As mentioned before, all domain categories are small categories and are therefore objects in the category $\cat$ of small categories. This category has all binary products and coproducts, so every pair of domain categories has products and coproducts.

\begin{figure}
    \centering
    \begin{subfigure}{0.45\textwidth}
\[\begin{tikzcd}[ampersand replacement=\&, sep=scriptsize]
	{\mathcal{Z}} \\
	\\
	\&\& {\C\times \D} \&\& \D \\
	\\
	\&\& \C
	\arrow["U", dashed, from=1-1, to=3-3]
	\arrow["G", from=1-1, to=3-5]
	\arrow["F"', from=1-1, to=5-3]
	\arrow["{\pi_2}"', from=3-3, to=3-5]
	\arrow["{\pi_1}", from=3-3, to=5-3]
\end{tikzcd}\]
    \caption{Product}
    \label{fig:cat-product}
    \end{subfigure}
    \begin{subfigure}{0.45\textwidth}
\[\begin{tikzcd}[ampersand replacement=\&,sep=scriptsize]
	\&\& \D \\
	\\
	\C \&\& {\C+ \D} \\
	\\
	\&\&\&\& {\mathcal{Z}}
	\arrow["{\phi_2}"', from=1-3, to=3-3]
	\arrow["G", from=1-3, to=5-5]
	\arrow["{\phi_1}", from=3-1, to=3-3]
	\arrow["F"', from=3-1, to=5-5]
	\arrow["U"', dashed, from=3-3, to=5-5]
\end{tikzcd}\]
        
    \caption{Coproduct}
    \label{fig:cat-coproduct}
    \end{subfigure}
    \caption{(a) Diagram showing the product of two categories $\C$ and $\D$ and the respective projections. (b) Diagram showing the coproduct of two categories $\C$ and $\D$ and the maps $\phi_1$ and $\phi_2$ mapping each category to the coproduct category.}
\end{figure}
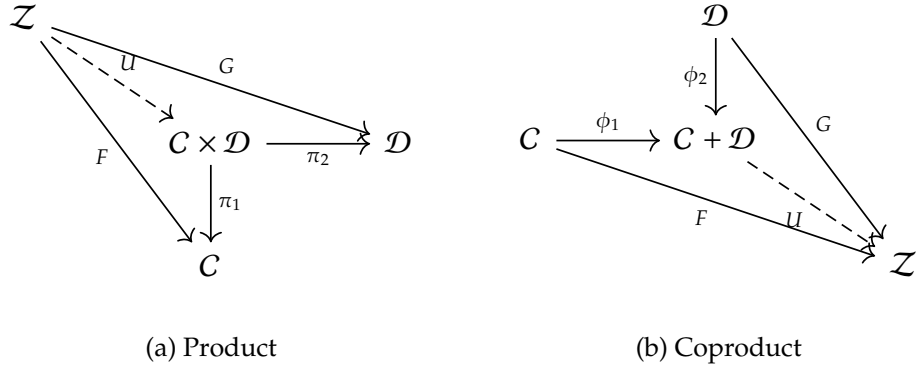

We can construct such a product $\C\times \D$ of two categories $\C$ and $\D$, that adheres to the properties of a product category as defined in Definition \ref{def:product} and shown in \autoref{fig:cat-product} as follows. The objects are $\Ob(\C\times\D) = \{(X,Y)\,|\,X\in\Ob(\C)$ and $Y\in\Ob(\D)\}$. For any two objects $(X,Y),(X',Y')\in\Ob(\C\times\D)$ the set of morphisms between them is defined as $\Hom_{\C\times\D}((X,Y),(X',Y'))=\{f\times g\,|\,f\in \Hom_\C(X,X')$ and $g\in \Hom_\D(Y,Y')\}$. Looking at two domains, this product construction can be seen as the overlay of all possible matches at once, including those that we excluded earlier, where objects with different names and functions are mapped.

There is a second way to combine two objects in a category or two categories in $\cat$.
A coproduct is the dual notion of a product as defined in Definition \ref{def:product}.
\begin{definition}[Coproduct]\label{def:coproduct}
    Let $\C$ be a category. A coproduct of two objects $X,Y\in \Ob(\C)$ is an object $X + Y\in \Ob(\C)$ with two projection morphisms $\phi_1\colon X\rightarrow X+ Y$ and $\phi_2\colon Y\rightarrow X+ Y$ such that for all objects $Z\in \Ob(\C)$ and morphisms $f\colon X\rightarrow Z$ and $g\colon Y\rightarrow Z$ there exists a unique $u\colon X+ Y\rightarrow Z$ such that $u\circ \phi_1 = f$ and $u\circ\phi_2 = g$. 
\end{definition}

The coproduct $\C + \D$ of two categories as shown in \autoref{fig:cat-coproduct} is just a disjoint union of the categories with the object set being the disjoint union of the sets of objects of the two categories, that is $\Ob(\C + \D) = \Ob(\C) \bigcup \Ob(\D)$. The morphisms only exist between objects from the same category, hence 
\begin{equation*}
    \Hom_{\C+\D}(X,Y) = 
    \begin{cases}
      \Hom_\C(X,Y), & \text{if}\ X,Y\in\Ob(\C) \\
      \Hom_\D(X,Y), & \text{if}\ X,Y\in\Ob(\D)\\
      \emptyset & \text{otherwise.}
    \end{cases}
  \end{equation*}

\begin{figure}[h]
    \centering
\begin{subfigure}{0.45\textwidth}
\begin{tikzcd}[ampersand replacement=\&, sep=scriptsize]
	Z \\
	\\
	\&\& {X\times_MY} \&\& Y \\
	\\
	\&\& X \&\& M
	\arrow["m_1", from=5-3, to=5-5]
	\arrow["m_2"', from=3-5, to=5-5]
	\arrow["{\pi_2}"', from=3-3, to=3-5]
	\arrow["{\pi_1}", from=3-3, to=5-3]
	\arrow["{f}"', from=1-1, to=5-3]
	\arrow["{g}", from=1-1, to=3-5]
	\arrow["\lrcorner"{anchor=center, pos=0.125}, draw=none, from=3-3, to=5-5]
	\arrow["u", dashed, from=1-1, to=3-3]
\end{tikzcd}
\caption{Pullback}
\label{fig:pullback}
\end{subfigure}
\begin{subfigure}{0.45\textwidth}
\begin{tikzcd}[ampersand replacement=\&, sep=scriptsize]
	M \&\& Y \\
	\\
	X \&\& {X+_MY} \\
	\\
	\&\&\&\& Z
	\arrow["m_2"', from=1-1, to=1-3]
	\arrow["m_1", from=1-1, to=3-1]
	\arrow["{\phi_2}"', from=1-3, to=3-3]
	\arrow["{g}", from=1-3, to=5-5]
	\arrow["{\phi_1}", from=3-1, to=3-3]
	\arrow["{f}"', from=3-1, to=5-5]
	\arrow["u", dashed, from=3-3, to=5-5]
\end{tikzcd}
\caption{Pushout}
\label{fig:pushout}
\end{subfigure}
\begin{subfigure}{0.45\textwidth}
\[\begin{tikzcd}[ampersand replacement=\&, sep=scriptsize]
	{\mathcal{Z}} \\
	\\
	\&\& {\C\times_\M \D} \&\& \D \\
	\\
	\&\& \C \&\& \M
	\arrow["U", dashed, from=1-1, to=3-3]
	\arrow["G", from=1-1, to=3-5]
	\arrow["F"', from=1-1, to=5-3]
	\arrow["{\Pi_2}"', from=3-3, to=3-5]
	\arrow["{\Pi_1}", from=3-3, to=5-3]
	\arrow["M_2"', from=3-5, to=5-5]
	\arrow["M_1", from=5-3, to=5-5]
\end{tikzcd}\]
\caption{Pullback in $\cat$}
\label{fig:cat-pullback}
\end{subfigure}
\begin{subfigure}{0.45\textwidth}
\[\begin{tikzcd}[ampersand replacement=\&, sep=scriptsize]
	\M \&\& \D \\
	\\
	\C \&\& {\C+_\M \D} \\
	\\
	\&\&\&\& {\mathcal{Z}}
	\arrow["M_2"', from=1-1, to=1-3]
	\arrow["M_1", from=1-1, to=3-1]
	\arrow["{\Phi_2}"', from=1-3, to=3-3]
	\arrow["G", from=1-3, to=5-5]
	\arrow["{\Phi_1}", from=3-1, to=3-3]
	\arrow["F"', from=3-1, to=5-5]
	\arrow["U"', dashed, from=3-3, to=5-5]
\end{tikzcd}\]

\caption{Pushout in $\cat$}
\label{fig:cat-pushout}
\end{subfigure}
    \caption{(a) Pullback: The pullback $X\times_M Y$ along $m_1$ and $m_2$ has a unique morphism $u$ for every $Z$ such that the diagram commutes. (b) Pushout: The pushout $X+_M Y$ along $m_1$ and $m_2$ has a unique morphism $u$ for every $Z$ such that the diagram commutes. (c) Pullback in $\cat$: The pullback of two categories $\C$ and $\D$ along functors to the middle category $\M$. (d) Pushout in $\cat$: The pushout of two categories $\C$ and $\D$ along functors from the middle category $\M$.}
\end{figure}
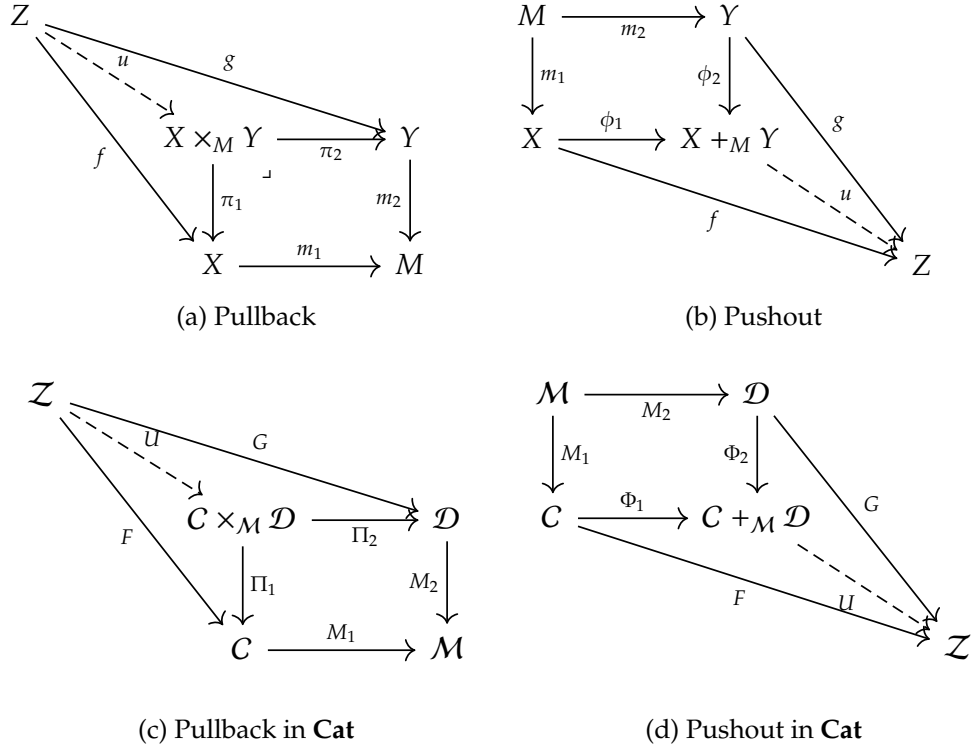

So far, these two concepts do not yield a nice analogy representation or blending, so we will look at the related concepts pullback and pushout to refine these constructions. A pullback can be understood as a restricted product (see \autoref{fig:pullback}).

\begin{definition}[Pullback]
A pullback $X\times_MY \in \Ob(\C)$ of two morphisms  $m_1\colon  X \rightarrow M$ and $m_2\colon Y\rightarrow M$ is equipped with the morphisms $\pi_1\colon X\times_MY\rightarrow X$ and $\pi_2\colon X\times_MY \rightarrow Y$ such that $m_1\circ\pi_1 = m_2\circ \pi_2$ and for any other object $Z\in \Ob(\C)$ and morphisms $f\colon Z\rightarrow X$ and $g\colon Z\rightarrow Y$ with $m_1\circ f = m_2\circ g$ there exists a unique morphism $u\colon Z \rightarrow X\times_MY$ such that $\pi_1\circ u = f$ and $\pi_2\circ u = g$. 
\end{definition}

The category $\cat$ of small categories has products and pullbacks. Pullbacks are unique up to isomorphisms, so renaming objects and morphisms of a pullback category does not change that it is a pullback. One such pullback of two categories $\C$ and $\D$ and two functors $M_1\colon \C\rightarrow\M$ and $M_2\colon \D\rightarrow\M$ as shown in \autoref{fig:cat-pullback} is the category $\C\times_\M\D$ with objects $\Ob(\C\times_\M\D)=\{(C,D)\,|\,C\in\Ob(\C), D\in\Ob(\D)$ and $M_1(C) = M_2(D)\}$, so for all objects in $\C$ and $\D$ that are mapped to the same object in $\M$ by $M_1$ and $M_2$, there is an object $(C,D)$ in $\C\times_\M\D$. For two objects $(C_1,D_1),(C_2,D_2)\in \C\times_\M\D$ the morphism set is $\Hom((C_1,D_1),(C_2,D_2))=\{(f,g)\,|\,f\in \Hom(C_1,C_2), g\in\Hom(D_1,D_2)$ and $M_1(f) = M_2(g)\}$, meaning the set contains one morphism for every pair of morphisms $f$ in $\C$ and $g$ in $\D$ that are mapped to the same morphism in $\M$ by $M_1$ and $M_2$. 

Let $F\colon \C\rightarrow\D$ be a functor describing a $\C$-$\D$ analogy. We can construct a pullback of $F\colon \C\rightarrow\D$ and $Id_\D\colon \D\rightarrow\D$ as shown in \autoref{fig:pullback_F}. Following the definition from above this is a new category $\C\times_F\D$ with $\Ob(\C\times_F\D) = \{(C,D)\,|\,C\in\Ob(\C), D\in\Ob(\D)$ and $F(C) = id_\D(D) = D\}$ and morphism sets $\Hom((C_1,D_1),(C_2,D_2))=\{(f,g)\,|\,f\in \Hom(C_1,C_2), g\in\Hom(D_1,D_2)$ and $F(f) = Id_\D(g) = g\}$. 

We can rename the objects, as most of them are pairs of similarly named objects (so we get $B_{\C,\D}$ instead of $(B_\C,B_\D)$ etc.). The resulting category contains all objects and morphisms that are mapped by $F$ and the information of what they are mapped to. We call this category the \emph{core} of the analogy $F$.

This core is a more abstract category in the sense that its objects do not need to be sets. All important information is in the morphisms. So, for example the object $B_{\C,\D}\in \Ob(\C\times_F\D)$ is just an abstract object representing the combination of $B_\C$ and $B_\D$. The morphisms \emph{true}$\colon 1_{\C,\D}\rightarrow B_{\C,\D}$ and \emph{false}$\colon 1_{\C,\D}\rightarrow B_{\C,\D}$ contain the information that this abstract domain contains a concept of true and false. There is an abstract object for every object match in the analogy and one combined morphism for every morphism match in the analogy. Hence, the core is a category describing exactly the matches in $F$.

The pushout (see \autoref{fig:pushout}) is the dual notion of pullback and can be seen as a refinement of the coproduct. The two categories are not combined disjointly but overlap in accordance to the additional object $M$ and morphisms $m_1$ and $m_2$.

\begin{definition}[Pushout]
    A pushout $X+_MY \in \Ob(\C)$ of two morphisms  $m_1\colon  M \rightarrow X$ and $m_2\colon M\rightarrow Y$ is equipped with the morphisms $\pi_1\colon X\rightarrow X+_MY$ and $\pi_2\colon Y\rightarrow X+_MY$ such that $\pi_1\circ m_1 =  \pi_2\circ m_2$ and for any other object $Z\in \Ob(\C)$ and morphisms $f\colon X\rightarrow Z$ and $g\colon Y\rightarrow Z$ with $ f\circ m_1 = g\circ m_2$ there exists a unique morphism $u\colon X+_MY \rightarrow Z$ such that $u\circ \pi_1= f$ and $u \circ \pi_2 = g$. 
\end{definition}

A pushout of two categories $\C$ and $\D$ and two functors $M_1\colon \M\rightarrow\C$ and $M_2\colon \M\rightarrow\D$ as shown in \autoref{fig:cat-pushout} is a category $\C+_\M\D$ with objects $\Ob(\C+_\M\D) = \Ob(\M)\cup \left(\Ob(\C)\setminus M_1(\Ob(\M))\right) \cup \left(\Ob(\D)\setminus M_2(\Ob(\M))\right)$ and morphism sets equal to the union of morphisms in $\C$ and $\D$ with only one representative for the images of morphisms in $\M$.

Let $F\colon \C\rightarrow \D$ be again a functor describing a $\C$-$\D$ analogy. We can construct a pushout of $F\colon \C\rightarrow \D$ and $Id_\C\colon \C\rightarrow \C$ as in \autoref{fig:pushout_F}. Following the definition from above this is a new category $\C+_F\D$ where $\Ob(\C+_F\D)$ is the union of both object sets considering $F$ and all morphism sets $\Hom_{\C+_F\D}(X,Y)=\{f\,|\,f\in \Hom_\C(X,Y)$ or $f\in\Hom_\D(X,Y)\}$. This can be seen as blend of the two categories with all additional objects and morphisms from $\D$ that are not in the image of $F$ and all objects and morphisms in $\C$ that are mapped to the same objects or morphisms in $\D$ being represented by single objects or morphisms.

So for each functor describing an analogy, we can construct the core of the analogy, that is a category that only contains parts that are also matched by the functor, and the blend, that is a category that contains both domains but includes the overlap described by the functor. The pullback $\C\times_F\D$ is also a pullback for $\C$ and $\D$ over $\C+_F\D$ and $\C+_F\D$ is also a pushout for $\C$ and $\D$ over $\C\times_F\D$ as shown in \autoref{sec:appendix}.

These two categories are very similar to $\C$ and do not cover all possible analogies. The more interesting case is when $F'\colon \C'\rightarrow \D$ is only a partial functor. In this case a pullback and pushout can be constructed using $\C'$ and an embedding $Em_\C\colon \C'\rightarrow\C$ as shown in \autoref{fig:pullback_F'} and \autoref{fig:pushout_F'}. In this case $\C\times_{F'}\D$ and  $\C+_{F'}\D$ are constructed as described above, but are not the pullback and pushout of each other.
\begin{figure}[h]
    \centering
\begin{subfigure}{0.23\textwidth}
\[\begin{tikzcd}[ampersand replacement=\&]
	{\C\times_F\D} \& \C \\
	\D \& \D
	\arrow["{\Pi_1}"', from=1-1, to=1-2]
	\arrow["{\Pi_2}", from=1-1, to=2-1]
	\arrow["\lrcorner"{anchor=center, pos=0.125}, draw=none, from=1-1, to=2-2]
	\arrow["F", from=1-2, to=2-2]
	\arrow["{Id_\D}"', from=2-1, to=2-2]
\end{tikzcd}\]
\caption{}
\label{fig:pullback_F}
\end{subfigure}
\begin{subfigure}{0.23\textwidth}
\[\begin{tikzcd}[ampersand replacement=\&]
	\C \& \C \\
	\D \& {\C+_F\D}
	\arrow["{Id_\C}"', from=1-1, to=1-2]
	\arrow["F", from=1-1, to=2-1]
	\arrow["{\Phi_1}", from=1-2, to=2-2]
	\arrow["{\Phi_2}"', from=2-1, to=2-2]
\end{tikzcd}\]
\caption{}
\label{fig:pushout_F}
\end{subfigure}
\begin{subfigure}{0.23\textwidth}
\[\begin{tikzcd}[ampersand replacement=\&]
	{\C\times_{F'}\D} \& \C' \\
	\D \& \D
	\arrow["{\Pi_1}"', from=1-1, to=1-2]
	\arrow["{\Pi_2}", from=1-1, to=2-1]
	\arrow["\lrcorner"{anchor=center, pos=0.125}, draw=none, from=1-1, to=2-2]
	\arrow["F'", from=1-2, to=2-2]
	\arrow["{Id_\D}"', from=2-1, to=2-2]
\end{tikzcd}\]
\caption{}
\label{fig:pullback_F'}
\end{subfigure}
\begin{subfigure}{0.23\textwidth}
\[\begin{tikzcd}[ampersand replacement=\&]
	{\C'} \& \C \\
	\D \& {\C+_{F'}\D}
	\arrow["{Em_\C}"', hook, from=1-1, to=1-2]
	\arrow["F'", from=1-1, to=2-1]
	\arrow["{\Phi_1}", from=1-2, to=2-2]
	\arrow["{\Phi_2}"', from=2-1, to=2-2]
\end{tikzcd}\]
\caption{}
\label{fig:pushout_F'}
\end{subfigure}
    \caption{(a) Pullback of the functor $F$ to construct the core of an analogy. (b) Pushout of the functor $F$ to construct the blend. (c) Pullback of the partial functor $F'$ to construct the core of an analogy. (d) Pushout of the partial functor $F'$ and an embedding $Em_\C\colon \C'\rightarrow \C$ to construct the blend.}
\end{figure}

We will look at our categories $\sun$ and  $\atom$ and construct an example core and blend. Let $F'\colon \sun'\rightarrow \atom$ be a partial functor that maps all objects $X_\sun\in\Ob(\sun)$ to $X_\atom\in\Ob(\atom)$ and maps the morphisms also according to their names. The partial category $\sun'$ does not contain the morphism $\te{hot}(\1)$ and any compositions with it, so those are also not part of $F$. The morphisms $\te{venus}(\1)$ and $\te{mars}(\1)$ are both mapped to the morphism $\te{electron}(\1)$. 

The core of this analogy is a category with objects $X_{\sun,\atom}$ for all $X_\sun\in \Ob(\sun)$. It contains the morphisms $\te{venus-electron}(\1)$ and $\te{mars-electron}(\1)$ and their concatenations with other morphisms, but does not contain the morphism $\te{hot}(\1)$.

The blend $\sun+_{F'}\atom$ is a category that consists of the same objects as the core, because all objects were mapped by the functor. But it also contains the additional morphism $\te{hot}(\1)$ and only one morphism $\te{venus-mars-electron}(\1)$, so these two possible matches were also blended together.

\section{Discussion}\label{sec:discussion}
In this paper, we have formalized knowledge domains as categories. By using category theory we focused on different types of predicates and relations and the information they contain about a domain's structure. We showed how a domain category can be constructed, that contains relevant structure of the domain and is small enough to restrict possible analogies. Using types and the form of a category allows us to view a domain through its different predicates and their different roles within their domain. These roles put strong constraints on possible analogies between domains. The objectification via exponential objects constrains this role even more, because it constrains how predicates and relations act as part of higher-order predicates. 
We used functors to define analogies that are constrained by these domain structures and the requirement of being truth-preserving. Using such a functor or, for more flexibility, a partial functor, we constructed the core and blend of an analogy. 

These functors are not necessarily unique, depending on the structure of the base- and target-domain. We have also not given a complete description of how to build these domain categories, and do not claim that there is one unique way to do so using \ac{ct}. We have rather looked at the basic concepts of construction and the effects these basic rules have.  
An important future direction is, obviously, to implement such categories in a computer program and automatically generate possible analogy functors. Such a generator should also rank possible functors using, for example, the amount of transferred morphisms as measure for the quality of an analogy.

We have focused on domains with entities and Boolean properties. It might be interesting to look at probabilistic domains or multi-valued logics. All this requires is using different sets for $B$. Another generalization is to add more base types, for example for quantities.

Our category-theoretic view on domains and analogies closely aligns with the classic structure mapping approach: Analogy-making is guided by the structure that predicates and relations impose on a domain. Category theory allows us to give a precise mathematical definition of what we mean by the term \emph{analogy}. One can argue about whether analogies should preserve truth and exponentiation or about the cognitively most plausible representation for a domain category. However, one can not argue about the fact that we need a precise working definition of analogies if we want to develop a mathematical theory of analogy-making. Perhaps such a theory will also help us to better understand the nature of human intelligence.

\paragraph{Funding.} 
 This work was funded by the Hessian Ministry of Higher Education, Research, Science and the Arts and its LOEWE research priority program ‘WhiteBox’ [grant number LOEWE/ 2/13/519/03/06.001(0010)/77].

\paragraph{acknowledgments}We would like to thank Michelle Geisler and Malte Ott for their helpful feedback and fruitful discussions. We would also like to thank two anonymous reviwers for their helpful comments.

\appendix
\section{Pullback and Pushout Combination}\label{sec:appendix}
\begin{lemma}\label{pull-push}
    Let $X$ and $Y$ be objects in a category with pullbacks and pushouts and $f:X\rightarrow Y$ be a morphism. If the left square in the diagram below forms a pullback with $X\times_fY$ and the right square forms a pushout with $X+_fY$, then  $X\times_fY$ is a pullback of $X$ and $Y$ over $\phi_1$, $\phi_2$ and $X+_f Y$, and $X+_f Y$ is a pushout of $X$ and $Y$ over $\pi_1$, $\pi_2$ and $X\times_fY$.
\end{lemma}

\[\begin{tikzcd}[ampersand replacement=\&]
	{X\times_fY} \& X \& X \\
	Y \& Y \& {X+_fY}
	\arrow["{\pi_1}"', from=1-1, to=1-2]
	\arrow["{\pi_2}", from=1-1, to=2-1]
	\arrow["\lrcorner"{anchor=center, pos=0.125}, draw=none, from=1-1, to=2-2]
	\arrow["{id_X}"', from=1-2, to=1-3]
	\arrow["f", from=1-2, to=2-2]
	\arrow["{\phi_1}", from=1-3, to=2-3]
	\arrow["{id_Y}"', from=2-1, to=2-2]
	\arrow["{\phi_2}"', from=2-2, to=2-3]
\end{tikzcd}\]

\begin{proof}

\begin{enumerate}
    \item  $X\times_fY$ is a pullback of $X$ and $Y$ over $\phi_1$, $\phi_2$ and $X+_f Y$:\\
\[\begin{tikzcd}[ampersand replacement=\&]
	Q \\
	\& {X\times_fY} \& X \\
	\& Y \& {X+_fY}
	\arrow["{u'}"', dashed, from=1-1, to=2-2]
	\arrow["g"', curve={height=-12pt}, from=1-1, to=2-3]
	\arrow["h"', curve={height=12pt}, from=1-1, to=3-2]
	\arrow["{\pi_1}"', from=2-2, to=2-3]
	\arrow["{\pi_2}", from=2-2, to=3-2]
	\arrow["{\phi_1}", from=2-3, to=3-3]
	\arrow["{\phi_2}"', from=3-2, to=3-3]
\end{tikzcd}\]
    Let $Q$ be another object in the category and $g:Q\rightarrow X$ and $h:Q\rightarrow Y$ be morphisms such that $\phi_1\circ g = \phi_2\circ h$. We need to show that there is a unique morphism $u:Q\rightarrow X\times_fY$ such that $g = \pi_1\circ u$ and $h = \pi_2\circ  u$.\\
    The right square is a pushout, so there is a unique morphism $u'$ such that $u'\circ \phi_1 = f$ and $u'\circ \phi_2 = id_Y$ because $id_Y\circ f = f \circ id_X$ (see diagram below).    
\[\begin{tikzcd}[ampersand replacement=\&]
	{X\times_fY} \& X \& X \\
	Y \& Y \& {X+_fY} \\
	\&\&\& Y
	\arrow["{\pi_1}"', from=1-1, to=1-2]
	\arrow["{\pi_2}", from=1-1, to=2-1]
	\arrow["\lrcorner"{anchor=center, pos=0.125}, draw=none, from=1-1, to=2-2]
	\arrow["{id_X}"', from=1-2, to=1-3]
	\arrow["f", from=1-2, to=2-2]
	\arrow["{\phi_1}", from=1-3, to=2-3]
	\arrow["f"', curve={height=-12pt}, from=1-3, to=3-4]
	\arrow["{id_Y}"', from=2-1, to=2-2]
	\arrow["{\phi_2}"', from=2-2, to=2-3]
	\arrow["{id_Y}", curve={height=12pt}, from=2-2, to=3-4]
	\arrow["{u'}", dashed, from=2-3, to=3-4]
\end{tikzcd}\]
Hence, the following equivalence holds:
$$id_Y\circ h = u'\circ \phi_2\circ h= u'\circ \phi_1\circ g = f\circ g.$$
The left square is a pullback, so there exists a unique morphism $u:Q\rightarrow X\times_f Y$ such that $g = \pi_1\circ u$ and $h = \pi_2\circ  u$.
\item $X+_f Y$ is a pushout of $X$ and $Y$ over $X\times_fY$:\\
\[\begin{tikzcd}[ampersand replacement=\&]
	{X\times_fY} \& X \\
	Y \& {X+_fY} \\
	\&\& Q
	\arrow["{\pi_1}"', from=1-1, to=1-2]
	\arrow["{\pi_2}", from=1-1, to=2-1]
	\arrow["{\phi_1}", from=1-2, to=2-2]
	\arrow["g"', curve={height=-12pt}, from=1-2, to=3-3]
	\arrow["{\phi_2}"', from=2-1, to=2-2]
	\arrow["h", curve={height=12pt}, from=2-1, to=3-3]
	\arrow["u", dashed, from=2-2, to=3-3]
\end{tikzcd}\]
  Let $Q$ be another object in the category and $g:X\rightarrow Q$ and $h:Y\rightarrow Q$ be morphisms such that $g \circ \pi_1 = h\circ \pi_2$. We need to show that there is a unique morphism $u:X+_fY\rightarrow Q$ such that $g = u\circ \phi_1$ and $h = u\circ \phi_2$.\\
    The left square is a pullback, so there is a unique morphism $u'$ such that $\pi_x\circ u' = f$ and $\pi_2\circ u' = id_Y$ because $id_Y\circ f = f \circ id_X$ (see diagram below).
\[\begin{tikzcd}[ampersand replacement=\&]
	X \\
	\& {X\times_fY} \& X \& X \\
	\& Y \& Y \& {X+_fY}
	\arrow["{u'}"', dashed, from=1-1, to=2-2]
	\arrow["{id_X}"', curve={height=-12pt}, from=1-1, to=2-3]
	\arrow["f"', curve={height=12pt}, from=1-1, to=3-2]
	\arrow["{\pi_1}"', from=2-2, to=2-3]
	\arrow["{\pi_2}", from=2-2, to=3-2]
	\arrow["\lrcorner"{anchor=center, pos=0.125}, draw=none, from=2-2, to=3-3]
	\arrow["{id_X}"', from=2-3, to=2-4]
	\arrow["f", from=2-3, to=3-3]
	\arrow["{\phi_1}", from=2-4, to=3-4]
	\arrow["{id_Y}"', from=3-2, to=3-3]
	\arrow["{\phi_2}"', from=3-3, to=3-4]
\end{tikzcd}\]
Hence, the following equivalence holds:
$$h\circ f = h\circ \pi_2\circ u'=g\circ \pi_1\circ u' = g\circ id_X.$$
The right square is a pushout, so there exists a unique morphism $u:X+_f Y\rightarrow Q$ such that $g = u\circ \phi_1$ and $h = u\circ \phi_2$.
\end{enumerate}
\end{proof}

Unfortunately, Lemma \ref{pull-push} only holds in this very specific case.

\bibliographystyle{apalike}
\bibliography{references}

\end{document}